\documentclass{article} 
\usepackage{iclr2025_conference,times}

\usepackage{amsmath,amsfonts,bm}

\def\eqref#1{equation~\ref{#1}}

\def\1{\bm{1}}

\DeclareMathAlphabet{\mathsfit}{\encodingdefault}{\sfdefault}{m}{sl}
\SetMathAlphabet{\mathsfit}{bold}{\encodingdefault}{\sfdefault}{bx}{n}

\usepackage[utf8]{inputenc} 
\usepackage[T1]{fontenc}    
\usepackage{hyperref}       
\usepackage{url}            
\usepackage{booktabs}       
\usepackage{amsfonts}       
\usepackage{nicefrac}       
\usepackage{microtype}      
\usepackage{xcolor}         

\usepackage{tcolorbox}
\tcbuselibrary{listings, breakable, skins, theorems}
\usepackage{amssymb}
\usepackage{pgfplots}
\usepackage{times}
\usepackage{latexsym}
\usepackage{multirow}
\usepackage{threeparttable}
\usepackage{xspace}
\usepackage{bm}
\usepackage{inconsolata}
\usepackage{pifont}
\usepackage{graphicx}
\usepackage{amsmath}
\usepackage{wrapfig}
\usepackage{multicol}
\usepackage{makecell}
\usepackage{tabularx}
\usepackage[rightcaption]{sidecap}
\usepackage{siunitx}
\usepackage{algorithm}
\usepackage{algpseudocode}
\usepackage{subcaption}
\usepackage{arydshln}
\usepackage{soul}

\newtcolorbox[auto counter, number within=section]{promptbox}[2][]{%
colback=gray!10!white,
colframe=gray!60!gray,
fonttitle=\bfseries\sffamily,
title=Prompt~\thetcbcounter: #2,
rounded corners,
arc=1.3mm,
boxrule=0.5pt,
enhanced,
breakable,
listing only,
listing options={
    basicstyle=\ttfamily\bfseries\itshape\fontsize{5}{6},
    numbers=left,
    numberstyle=\tiny\color{gray!80!black},
    stepnumber=1,
    numbersep=5pt,
    showspaces=false,
    showstringspaces=false
},
label={prompt:#1}
}

\renewcommand{\raggedright}{\leftskip=0pt \rightskip=0pt plus 0cm}

\newcommand{\ie}{\emph{i.e.,}\xspace}
\newcommand{\eg}{\emph{e.g.,}\xspace}

\newcommand{\name}{\textsc{CtrlA}\xspace}
\definecolor{alizarin}{rgb}{0.82, 0.1, 0.26}
\definecolor{ballblue}{rgb}{0.13, 0.67, 0.8}

\title{\name: Adaptive Retrieval-Augmented Generation via Inherent Control}

\author{Huanshuo Liu$^{1,\gimel}$,\; Hao Zhang$^{1,\gimel,\varkappa}$,\; Zhijiang Guo$^{1,\varkappa}$,\; Jing Wang$^2$\\
\textbf{Kuicai Dong$^1$, Xiangyang Li$^1$,\; Yi Quan Lee$^1$,\; Cong Zhang$^1$,\; Yong Liu$^1$} \\
$^1$Noah's Ark Lab, Huawei Technologies Co., Ltd\\
$^2$ Individual Researcher
}

\iclrfinalcopy 
\begin{document}

\maketitle

\def\thefootnote{$\gimel$} \footnotetext{The first two authors contributed equally.}\def\thefootnote{\arabic{footnote}}
\def\thefootnote{$\varkappa$} \footnotetext{Correspondence to Hao Zhang\texttt{<hzhang26@outlook.com>} and Zhijiang Guo\texttt{<cartusguo@gmail.com>}.}\def\thefootnote{\arabic{footnote}}

\begin{abstract}
Retrieval-augmented generation (RAG) has emerged as a promising solution for mitigating hallucinations of large language models (LLMs) with retrieved external knowledge. Adaptive RAG enhances this approach by enabling dynamic retrieval during generation, activating retrieval only when the query exceeds LLM's internal knowledge. Existing methods primarily focus on detecting LLM's confidence via statistical uncertainty. Instead, we present the first attempts to solve adaptive RAG from a representation perspective and develop an inherent control-based framework, termed \name. Specifically, we extract the features that represent the honesty and confidence directions of LLM and adopt them to control LLM behavior and guide retrieval timing decisions. We also design a simple yet effective query formulation strategy to support adaptive retrieval. Experiments show that \name is superior to existing adaptive RAG methods on a diverse set of tasks, the honesty steering can effectively make LLMs more honest and confidence monitoring is a promising indicator of retrieval trigger.\footnote{Our code is available at \url{https://github.com/HSLiu-Initial/CtrlA}}
\end{abstract}

\section{Introduction}\label{sec:intro}
Retrieval-augmented generation (RAG;~\citealt{guu2020realm,izacard2022atlas}) has proven effective in mitigating hallucination by integrating external knowledge into LLMs. Early efforts often employ single-round, indiscriminate retrieval, resulting in over-reliance on external knowledge and incomplete retrieval~\citep{wang2023selfknowledge,su2024bright}. To solve the issues, adaptive RAG (ARAG;~\citealt{jiang2023active,wang2024selfdc}) has emerged, which enables dynamic retrieval during generation, activating retrieval only when the query exceeds LLM's internal knowledge~\citep{ni2024llms}. 

The key challenges in ARAG involve determining \textit{what} and \textit{when} to retrieve~\citep{su2024dragin,yao2024seakr}. The design of \textit{what} aspect typically depends on the construction of \textit{when} aspect, making ARAG’s primary focus the issues related to \textit{when} aspect. For the \textit{when} aspect, recent ARAGs leverage the ability that LLMs are aware of their uncertainty~\citep{kuhn2023semantic,chen2024inside,xiong2024can}, utilizing this characteristic to determine retrieval timing by assessing \textit{confidence} level of their knowledge~\citep{su2024dragin,yao2024seakr}. They primarily focus on detecting uncertainty in the LLM's outputs to signal retrieval, relying on factors such as output probabilities~\citep{jiang2023active}, entropy of output~\citep{su2024dragin} or internal states~\citep{yao2024seakr}, or verbal feedback~\citep{wang2024llms,yan2024corrective}. 
From a statistical standpoint, uncertainty and confidence are conceptually equivalent, both reflect the degree of certainty in a model's predictions~\citep{yang2023improving,band24linguistic,tao2024trust}. Thus, uncertainty can serve as a proxy for confidence in determining retrieval timing.

We revisit the assumptions underlying these uncertainty-based methods. First, they presume that LLM's output aligns with its internal knowledge~\citep{lin2022truthfulqa,zou2023representation}, that is, LLM can accurately reflect its internal knowledge in outputs, \ie they are \textit{honest}. However, LLMs often navigate a trade-off between honesty and helpfulness, balancing discerning its limitations and generating user-satisfied plausible content~\citep{liu2024how}. When the output diverges from internal knowledge, indicating low honesty, they only detect intended output rather than internal knowledge. Second, they equate uncertainty with LLM's \textit{confidence},~\footnote{Confidence is the feeling of belief or trust that a person or thing is reliable~\citep{Bandura1977SelfEfficacyTE}.} which may be not always applicable to LLM behavior. For instance, an LLM may frequently respond with ``I don't know'' or ``insufficient information,'' suggesting low uncertainty, yet retrieval should still occur. Moreover, semantically equivalent answers can be expressed in various ways in free-form generation, which may leads to high uncertainty~\citep{farquhar2024detecting}. However, retrieval is unnecessary in this scenario.

Based on this analysis, we emphasize both \textit{honesty} and \textit{confidence} of LLMs are crucial for accurate retrieval timing. However, current ARAGs struggle to address them due to the limitations of statistical uncertainty. 
We propose to solve ARAG from a representation perspective~\citep{olah2023distributed,bricken2023towards,zou2023representation,templeton2024scaling}, developing an efficient and unified framework that seamlessly tackles the requirements of honesty and confidence. Our core idea involves extracting features corresponding to \textit{honesty} and \textit{confidence} directions from LLMs and using them to control LLM behavior and guide retrieval timing decisions simultaneously. 

We devise an Inherent \textbf{C}on\textbf{tr}o\textbf{l}-based \textbf{A}daptive RAG framework (\textbf{\name}). To steer LLM toward honesty and monitor its confidence, we extract features aligned with the directions of honesty and confidence within LLM's representation space. By adjusting the honesty direction—a process we refer to as \textcolor{alizarin}{\textit{\textbf{honesty steering}}}—we can shift the LLM's representation space to promote more honesty outputs. Simultaneously, confidence is quantified by measuring the projection of current representation onto the confidence feature, a method we call \textcolor{ballblue}{\textit{\textbf{confidence monitoring}}}. Honesty steering helps LLM recognize its limitations and suppress the generation of fabricated plausible information. Confidence monitoring, in turn, enhances the precision of retrieval timing. We also implement a simple yet effective query formulation module to support adaptive retrieval, minimizing the impact of noise and intent drift. Extensive experiments verify the effectiveness of \name, revealing that adjusting the directions of LLM's internal states enhances its honesty, while confidence monitoring reliably signals when to trigger retrieval, optimizing the balance between retrieval and internal knowledge use.

\section{Related Work}\label{sec:related_work}

\subsection{Retrieval-Augmented Generation}\label{ssec:rag}
Early RAG efforts~\citep{lewis2020retrieval, karpukhin2020dense, zhu2021retrieving, komeili2022internet, khattab2023demonstrate,liu2024retrieval} relied on single-round, indiscriminate retrieval, increasing computational costs and degrading model performance~\citep{wang2023selfknowledge, su2024bright,zhang2024evaluating}. To address these issues, ARAG emerged, enabling dynamic retrieval during generation when the query exceeds the LLM's internal knowledge~\citep{jiang2023active, wang2024selfdc, ni2024llms}. Previous implementations utilized static rules, such as prior sentences~\citep{trivedi2023interleaving}, sliding windows~\citep{borgeaud2022improving, ram2023context}, and in-context learning~\citep{zhao2023verify, zhang2024retrievalqa, li2024chainofknowledge}. Recent ARAGs leverage LLMs' self-awareness of uncertainty to optimize retrieval timing by assessing confidence levels through internal states~\citep{yao2024seakr}, likelihoods~\citep{jiang2023active, wang2024selfdc, su2024dragin}, or verbal feedback~\citep{wang2024llms, ding2024rowen, yan2024corrective}. This enhances retrieval timing and balances external and internal knowledge use. However, uncertainty-based ARAGs face challenges with LLM honesty and confidence, crucial for accurate retrieval timing. \name addresses these issues from a representation perspective, enhancing control over honesty and confidence to improve retrieval-augmented generation effectiveness.

\subsection{Linear Representations in LLMs}\label{ssec:rep}
Recent research has explored LLM representations to understand their beliefs, interpretability, and compliance~\citep{levinstein2023still,li2023emergent,bricken2023towards}. Grounded in the linear representation and superposition hypotheses, these studies suggest that specific features can be aligned with particular directions in the LLMs' linear space. This framework effectively guides and monitors model outputs~\citep{olah2023distributed}. Researchers have modified or detected models' demeanor, preferences, stated goals, and biases, as well as induced errors or mitigated risks~\citep{templeton2024scaling}. Supporting the hypotheses, \citet{marks2023thego} and \citet{slobodkin2023curious} found that features like truthfulness and answerability are linearly separable within the latent space. Further efforts~\citep{zou2023representation,liu2023aligning} utilized contrastive instruction templates to clarify feature directions.  \name leverages these insights to extract features related to honesty and confidence, aiming to control LLM behavior and guide retrieval timing decisions, bridging representational understanding and practical applications.
\section{Inherent Control based Adaptive RAG}\label{sec:method}

\subsection{Preliminary}\label{ssec:preliminary}
Given a query $\bm{q}$, RAG aims to assist LLMs in generating more precise answers $\bm{y}=[s_1,\dots,s_m]=[w_1,\dots,w_n]$ containing $m$ sentences or $n$ tokens by retrieving relevant documents $\mathcal{D}_q=\mathcal{R}(\bm{q})$ from document corpus $\mathcal{D}=\{\bm{d}_i\}_{i=1}^{|\mathcal{D}|}$ or web via retriever $\mathcal{R}$. The retrieved documents $\mathcal{D}_q$ are usually concatenated with input $\bm{x}$, \ie query $\bm{q}$ with task instruction $\mathcal{I}$, to aid answer generation as $\bm{y}=\mathtt{LLM}([\mathcal{D}_q;\bm{x}])$, where $[\cdot;\cdot]$ denotes concatenation. 
In contrast, adaptive RAG performs active retrieval necessity decision via a trigger mechanism $\mathcal{T}(\bm{x},\bm{y}_{<t})$, where $\bm{y}_{<t}$ is the output segment as of step $t (t\geq 1)$. If $\mathcal{T}$ is triggered, the query formulation function $\bm{q}_t=f_q(\bm{x},\bm{y}_{<t})$ will produce a query $\bm{q}_t$ to search. If $\mathcal{T}$ is triggered at $t=1$, \ie $\bm{y}_{<1}=\emptyset$, $\bm{q}$  will be the original query. Given the retrieved documents $\mathcal{D}_{q_t}$, the model continues generating the next output segment $\bm{y}_t=\mathtt{LLM}([\mathcal{D}_{q_t};\bm{x};\bm{y}_{<t}])$ till the answer comes to its end or next retrieval trigger occurs.

\begin{figure}[t]
    \centering
    \includegraphics[trim={0cm 0.2cm 0.4cm 0.0cm},clip,width=\textwidth]{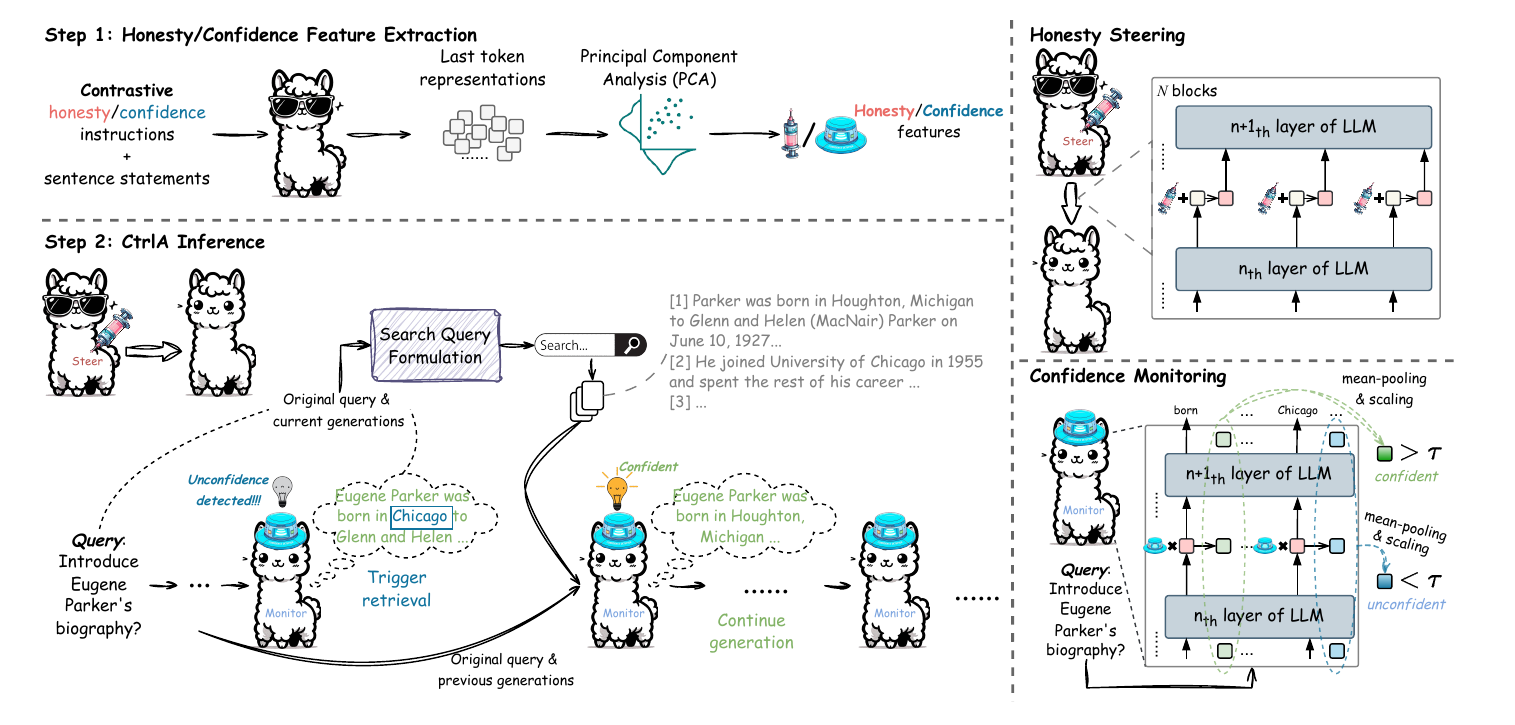}
    \captionsetup{skip=0pt}
    \caption{\name framework. Step 1 extracts the features corresponding to \textcolor{alizarin}{\textit{honesty}} and \textcolor{ballblue}{\textit{confidence}} directions; Step 2 utilizes extracted features to steer and monitor LLM behaviors at inference. The \textcolor{alizarin}{\textit{honesty}} feature \textbf{steers} the representation of LLM to make it more honest, while \textcolor{ballblue}{\textit{confidence}} feature is used to \textbf{monitor} the confidence level of LLM outputs, where the token whose score is lower than the threshold is marked as unconfident. The retrieval is triggered if specific tokens are unconfident.}
    \label{fig:framework}
    \vspace{-0.4cm}
\end{figure}

\subsection{\name Framework}\label{ssec:ctrla}

\subsubsection{Linear Representation Feature Extraction}\label{sssec:monosemantic}
Our approach builds on the linear representation and superposition hypotheses~\citep{olah2023distributed,bricken2023towards,templeton2024scaling}. We aim to extract features that represent \textit{honesty} and \textit{confidence} directions from LLM's representation space and use them to steer or monitor its behavior. Specifically, we manually craft contrastive instructions, as shown in Prompt~\ref{prompt:exp_ref_template}, to extract features that represent the directions of honesty and confidence. Let $\mathcal{I}_{h/c}^+$ denote the positive instruction of honest or confident, $\mathcal{I}_{h/c}^-$ be the negative instruction of dishonest or unconfident, and $\mathcal{S} = \{ \bm{s}_1, \dots, \bm{s}_{|\mathcal{S}|} \}$ represent the dataset with $|\mathcal{S}|$ statements used for extract target features. 

For honesty feature extraction, each statement $s_i$ is concatenated with both positive and negative instructions, forming $\mathcal{I}_h^+ \oplus s_i$ and $\mathcal{I}_h^- \oplus s_i$, respectively, resulting in $|\mathcal{S}|$ statement pairs. For the statement pair of $s_i$, they are sequentially fed into LLM in a teacher-forcing manner to collect token representations. Given that each LLM layer encodes a unique semantic space~\citep{chuang2023dola,sun2024transformer}, we extract token representations from all LLM layers. Assuming LLM has $L$ layers and $\bm{s}_i$ contains $n$ tokens, we obtain representations  $\{\{\bm{r}_{i,k}^{l,+}\}_{k=1}^{n}\}_{l=1}^{L}$ and $\{\{\bm{r}_{i,k}^{l,-}\}_{k=1}^{n}\}_{l=1}^{L}$ for positive and negative instructions, where $\bm{r}_{i,k}^{l}$ denotes the $k$-th token representation of $s_i$ at layer $l$. The contrastive vector for the $k$-th token at $l$-th layer is computed as $\bm{v}_{i,k}^l=\bm{r}_{i,k}^{l,+}-\bm{r}_{i,k}^{l,-}$. Thus, after processing all tokens of $s_i$, we derive the set of contrastive vectors $\{\{\bm{v}_{i,k}^{l}\}_{k=1}^{n}\}_{l=1}^{L}$. Since we employ \textit{teacher-forcing} to encode each token's representation—and ``honest'' and ``dishonest'' are the only differing descriptions between $\mathcal{I}_h^+$ and $\mathcal{I}_h^-$—the vector $\bm{v}_{i,k}^{l}$ captures the honesty direction for the $k$-th token at layer $l$~\citep{zou2023representation}. After processing all statements in $\mathcal{S}$, we apply PCA to the collected contrastive vectors $\{\{\{\bm{v}_{i,k}^{l}\}_{k=1}^{n}\}_{l=1}^{L}\}_{i=1}^{|\mathcal{S}|}$ at each layer $l$, extracting the first principal component as the general honesty direction. This results in a set of honesty direction vectors $\bm{v}_h=\{\bm{v}_{h}^{l}\}_{l=1}^{L}$. Note confidence feature extraction also utilizes the same method to derive $\bm{v}_c=\{\bm{v}_{c}^{l}\}_{l=1}^{L}$.

\begin{promptbox}[exp_ref_template]{Instruction for Honesty and Confidence Feature Extraction}
\textbf{[INST]} \texttt{Pretend you're a <honest/dishonest> | <confident/unconfident> person making statements about the world.} \textbf{[/INST]} \texttt{<a statement $s_i$>}
\end{promptbox}

We use the True-False dataset~\citep{azaria2023internal} as $\mathcal{S}$ for honesty feature extraction, which tests whether LLMs' internal states reflect truthfulness. For confidence, we synthesize confident and unconfident statements using GPT-4 (ref. Appendix~\ref{appd:ssec:probe_dataset}) due to the scarcity of available datasets.

\subsubsection{Honesty Steering}\label{sssec:honesty_steer}
According to the superposition hypothesis, adjusting LLM by moving each token's representation closer to the direction representing the honesty feature during decoding, is a direct way to enhance its honesty~\citep{olah2023distributed,zou2023representation,templeton2024scaling}. To achieve this, we employ a simple linear combination. After extracting the honesty feature, it can be directly used to steer the behavior of the LLM. Assuming the LLM contains $L$ layers, each layer has its corresponding feature. Let $\bm{v}_h = \{\bm{v}_{h}^l\}_{l=1}^{L}$ denote the honesty feature and $\bm{R}_k = \{\bm{r}_{k}^l\}_{l=1}^{L}$ represent the token representations for the $k$-th token at each layer. We then apply a linear combination function for honesty steering:
\begin{equation}
    \hat{\bm{R}}_k=\bm{R}_k+\lambda\cdot\bm{v}_h=\{\bm{r}_{k}^l+\lambda\cdot\bm{v}_{h}^l \; | \; \forall \; l \in [1,\dots,L]\},
    \label{eq:honesty_control}
\end{equation}
where the coefficient $\lambda$ controls the strength of honesty steering. Because $\bm{v}_h$ represents the direction that promotes honesty, the ``$+$'' operator is used in Eq~\ref{eq:honesty_control}. Conversely, to reduce honesty, the ``$-$'' operator can be employed. As illustrated in Figure~\ref{fig:framework}, honesty steering is applied layer-by-layer and token-by-token during generation. This method is both simple and effective, with minimal impact on inference costs. For brevity, we denote honesty steering as $\hat{\bm{y}}_t = \mathcal{P}_h(\bm{y}_t)$ in the following descriptions.

\subsubsection{Confidence Monitoring as Retrieval Trigger}\label{sssec:conf_monitor}
According to the linear representation hypothesis, an intuitive way to monitor an LLM's confidence during generation is to evaluate how well token representations align with the confidence feature direction in the representation space~\citep{bricken2023towards,zou2023representation,templeton2024scaling}. Given the extracted confidence feature, we utilize it to monitor LLM's confidence during generation. Let $\bm{R}_k=\{\bm{r}_{k}^l\}_{l=1}^{L}$ represent the $k$-th token's representation at each layer, and $\bm{v}_c=\{\bm{v}_{c}^l\}_{l=1}^{L}$ denote the confidence feature. Specifically, we compute the confidence score for $k$-th token using the dot product, followed by mean-pooling across layers and a scaling operation for normalization and outlier removal. This produces the confidence score for the $k$-th token as follows:
\begin{equation}
\begin{aligned}
    \tilde{m}_k &= \mathtt{meanpool}([m_{k,1}, \dots, m_{k,L}]) = \mathtt{meanpool}\big([\bm{r}_{k}^{l,\top}\cdot\bm{v}_{c}^l]_{l=1}^{L}\big), \\
    \bar{m}_k &= \mathtt{scale}([\tilde{m}_0,\dots,\tilde{m}_k])[-1] - \tau,
\end{aligned}
\end{equation}
where $\tau$ is the threshold to adjust the sensitivity of confidence monitoring, $\tilde{m}_{<k}$ represents the mean-pooled score of preceding tokens, and the index $-1$ refers to the score of the last token, \ie $k$-th token. If $\bar{m}_k > 0$, it suggests that the $k$-th token's representational direction leans towards the confidence, indicating that LLM is confident in generating this token. Conversely, if $\bar{m}_k < 0$, LLM is unconfident in generating the $k$-th token. Here we denote confidence monitoring as $\mathcal{P}_c$.

For the $t$-th output segment $\hat{\bm{y}}_t=[w_{t_s},\dots,w_{t_e}]$ of the LLM, with confidence scores $[\bar{m}_{t_s},\dots,\bar{m}_{t_e}]$ for each token, the retrieval necessity is measured by the confidence scores of specific tokens within $\hat{\bm{y}}_t$. We only consider the confidence scores of \emph{new information} in $\hat{\bm{y}}_t'$, \ie content that has not appeared in the previous generation and excludes trivial tokens, like stopwords. The retrieval trigger $\mathcal{T}$ activates if any confidence score in $\hat{\bm{y}}_t'$ satisfies $\bar{m}_k < 0$, where $t_s \leq k \leq t_e$. If $\hat{\bm{y}}_t'$ contains such tokens, retrieval is triggered, \ie $\mathcal{T}(\mathcal{P}_c(\hat{\bm{y}}_t')) == \mathtt{True}$.

\subsubsection{Search Query Formulation}\label{sssec:qform}
Upon retrieval triggered, we need to employ a search query to retrieve relevant documents that aid in LLM generation. The construction of effective search queries plays a pivotal role in enhancing retrieval efficiency. We develop two search query formulation strategies.

\paragraph{Context-Augmented Querying.} Initially, for a query $\bm{q}$, we prompt the LLM to sequentially generate responses. Once the retrieval is triggered, context-augmented querying (CAQ) will concatenate the query $\bm{q}$ with the processed output segment $\hat{\bm{y}}_t$ for retrieval, since using the original query as a supplement can avoid intent drift and improve the effectiveness of retrieval~\citep{jagerman2023query}. Besides, the output segment $\hat{\bm{y}}_{t}=[w_{t_s}, \ldots, w_{t_e}]$ may contain noise such as unconfident tokens and incorrect contents, we process the sentence by masking out the tokens, which satisfy (i) not appeared in $\bm{q}$ and previous generations $\bm{y}_{<t}$, \ie new information and (ii) unconfident tokens, as:
\begin{equation}
\mathtt{mask}(\hat{\bm{y}}_t) = \left\{ \bar{w} \middle| \bar{w} =
\begin{cases}
\emptyset, & \text{if } w \not\in \bm{q} \cup \bm{y}_{<t} \text{ and } \bar{m}_w < 0\\
w, & \text{otherwise}
\end{cases}, \forall w \in \hat{\bm{y}}_t \right\}.
\label{eq:mask_function}
\end{equation}

\begin{wrapfigure}{R}{0.55\textwidth}
\vspace{-22pt}
    \begin{minipage}{0.55\textwidth}
      \begin{algorithm}[H] 
\caption{\name Inference}
\begin{algorithmic}[1]
\Require Language Model $\mathtt{LM}$, Retriever $\mathcal{R}$, Document Corpus $\mathcal{D}$, Honesty Steering $\mathcal{P}_h$, Query Formulator $f_q$, Retrieval Trigger $\mathcal{T}$
\State \textbf{Input:} input prompt $\bm{x}$ ($\mathcal{I}$ and $\bm{q}$), previous generation $\bm{y}_{<t}$
\State \textbf{Output:} next output segment $\bm{y}_t$

\State \texttt{LLM} along with $\mathcal{P}_h$ predicts next segment $\hat{\bm{y}}_t$ given $(\bm{x},\bm{y}_{<t})$

\State $\mathcal{T}$ simultaneously monitors retrieval signal during \texttt{LLM} generates $\hat{\bm{y}}_t$

\If{$\mathcal{T}\text{ == }\texttt{True}$}

    \State $\mathcal{R}$ retrieves $\mathcal{D}_q$ from $\mathcal{D}$ using $\bm{q}_t=f_q(\bm{q},\hat{\bm{y}}_{t})$
    
    \State \texttt{LM} along with $\mathcal{P}_h$ re-predicts next segment $\hat{\bm{y}}_t$ given $(\bm{x},\bm{y}_{<t}, \mathcal{D}_q)$

\EndIf
\State Set $\bm{y}_t=\hat{\bm{y}}_t$
\end{algorithmic}
\label{alg:overview_light}
\end{algorithm}
    \end{minipage}
    \vspace{-20pt}
  \end{wrapfigure}

Thus, the CAQ generates the refined search query as $f_{\text{CAQ}}(\bm{x},\hat{\bm{y}}_t)=\left[\bm{q};\mathtt{mask}(\hat{\bm{y}}_t)\right]$.

\paragraph{Targeted Validation Querying.} CAQ directly masks out the noise of the output segment and concatenates it with the original query to form a search query. Yet, off-the-shelf retrievers may prefer a well-formatted query~\citep{karpukhin2020dense}. Thus, we also develop a targeted validation querying strategy (TVQ), $f_{\text{TVQ}}$. It instructs LLM to produce search query using original query and current output segment as references (see Prompt~\ref{prompt:tvq}). The goal of TVQ is to generate a query to validate the accuracy of current output segment by searching for supporting documents. For simplicity, we use $f_q$ to represent both $f_{\text{CAQ}}$ and $f_{\text{TVQ}}$.

\subsection{Inference Process}\label{ssec:infer}
For an input $\bm{x}$ and preceding generation $\bm{y}_{<t}$, the model generates the output segment along with honesty steer $\mathcal{P}_h$ and derives $\hat{\bm{y}}_t$. Simultaneously, the confidence monitor $\mathcal{P}_c$ is activated to compute the confidence score of each token during generation. We collect the confidence scores of new information $\hat{\bm{y}}_t'$ to determine retrieval necessity via retrieval trigger $\mathcal{T}$. If retrieval is not required, the model continues predicting the next output segment. Otherwise, we adopt query formulation, $f_q$, to produce search query $\bm{q}_t$ and retrieve documents $\mathcal{D}_q$ via retriever $\mathcal{R}$. The retrieved documents $\mathcal{D}_q$, input $\bm{x}$, and preceding generation $\bm{y}_{<t}$ are concatenated to regenerate the current output segment. Algorithm~\ref{alg:overview_light} presents an overview of \name inference step. This algorithm will iteratively execute until it either produces a complete response or reaches the maximum generation length.

\section{Experiment Setup}\label{sec:exp_setup}
\textbf{Datasets and Evaluation.} For \textit{short-form} QA, we select PopQA~\citep{mallen2022not} and TriviaQA~\citep{joshi2017triviaqa}. For \textit{long-form} QA, we use ASQA~\citep{stelmakh2022asqa} and biography generation (Bio; \citealt{min2023factscore}). For \textit{multi-hop} QA, we follow \citet{su2024dragin} to choose 2WikiMultihopQA (2WMQA; \citealt{ho2020constructing}) and HotpotQA (HQA; \citealt{yang2018hotpotqa}). For short-form QA, we report the accuracy. For ASQA, we report str-em, Rouge-L (R-L; \citealt{lin2004rouge}), MAUVE (mau; \citealt{pillutla2023mauve}), EM and F1. Bio is evaluated by FactScore (FS; \citealt{min2023factscore}). For multi-hop QA, we report EM and F1. We also evaluate $500$ test samples (\texttt{v04082024}) of FreshQA~\citep{vu2023freshllms} and report relaxed and strict accuracy scores. More details in Appendix~\ref{apd:ssec:dataset} and~\ref{apd:ssec:metric}.

\textbf{Implementation and Retrieval Setup.} We select the Mistral-7B~\citep{jiang2023mistral} as the backbone of \name and adopt the greedy decoding strategy for all experiments. The $\lambda$ for honesty steer is set as $0.3$ and the $\tau$ for confidence monitoring is set as $0.0$. By default, we use BM25 and BGE~\citep{xiao2023c} as our retriever and use the 2018 English Wikipedia corpus as document source following \citet{jiang2023active} and \citet{asai2024selfrag}. For PopQA and Bio, we follow Self-RAG~\citep{asai2024selfrag} to additionally retrieve from the web to mitigate the coverage limitations in the Wikipedia corpus. For the multi-hop QA task, we only use BM25 as the retriever. For FreshQA, we only retrieve from the web to obtain supporting documents. More details in Appendix~\ref{apd:ssec:impl} and~\ref{apd:ssec:retriever}.

\textbf{Baselines.} We compare \name with representative RAG baselines: (1) Single-round RAG (SR-RAG), which retrieves documents before generation; (2) Fix-sentence RAG (FS-RAG; \citealt{trivedi2023interleaving}), which triggers retrieval every sentence and the previous sentence is used as query; (3) Fix-length RAG (FL-RAG; \citealt{ram2023context}), which triggers retrieval every $n$ tokens and the previous token window is used as query; (4) Query-decompose RAG (QD-RAG; \citealt{press2023measuring,khattab2023demonstrate}), which prompts LLMs to generate follow-up queries and trigger retrieval for each query; (5) Adaptive RAGs: FLARE~\citep{jiang2023active}, Self-RAG~\citep{asai2024selfrag}, DRAGIN~\citep{su2024dragin}, SeaKR~\citep{yao2024seakr}, RQ-RAG~\citep{chan2024rqrag} and QC-RAG~\citep{jeong2024adaptive}. \textbf{For (1)-(4), we reimplement them under the same setting as \name}. More details about the baselines are in Appendix~\ref{apd:ssec:baseline}.

\section{Experiment Results and Analysis}\label{sec:results}

\subsection{Main Results}\label{ssec:main_result}

\begin{wraptable}{r}{0.4\textwidth}
    \centering
    \small
    \caption{Overall results of short-form QA. $^{\diamond}$ is our reproduced results. $^{\ddagger}$ denotes results in the corresponding work.}
    \vspace{-2mm}
    \begin{tabular}{l c c}
    \toprule
    Method & TriviaQA & PopQA \\
    \midrule
    wo-RAG$_{\text{7B}}^{\diamond}$ & 53.8 & 25.7 \\
    SR-RAG$_{\text{7B}}^{\diamond}$ & 62.7 & 51.9 \\  
    FL-RAG$_{\text{7B}}^{\diamond}$ & 60.8 & 28.1 \\
    FS-RAG$_{\text{7B}}^{\diamond}$ & 54.3 & 26.9 \\
    QD-RAG$_{\text{7B}}^{\diamond}$ & 52.3 & 29.4 \\
    \midrule
    FLARE$_{\text{7B}}^{\diamond}$ & \underline{72.4} & 48.3 \\
    Self-RAG$_{\text{7B}}^{\ddagger}$ & 66.4 & 54.9 \\
    Self-RAG$_{\text{13B}}^{\ddagger}$ & 69.3 & 55.8 \\
    RQ-RAG$_{\text{7B}}^{\ddagger}$ & - & \underline{57.1} \\
    QC-RAG$_{\text{11B}}^{\ddagger}$ & 58.2 & - \\
    \textbf{\name}$_{\text{7B}}$ & \textbf{76.4} & \textbf{61.8} \\
    \bottomrule
    \end{tabular}
    \label{tab:short_form_results}
    \vspace{-2mm}
\end{wraptable}

\textbf{Performance comparison.}
\name demonstrates consistent superiority over the compared approaches across various tasks and evaluation metrics, as evidenced by the results in short-form QA (Table~\ref{tab:short_form_results}), long-form QA (Table~\ref{tab:long_form}), multi-hop QA (Table~\ref{tab:multihop_results}), and the FreshQA dataset (Table~\ref{tab:freshqa_results}). In each case, \name surpasses fine-tune based methods (\eg Self-RAG), uncertainty-based methods (\eg FLARE and DRAGIN), and rule-based methods (\eg FL/FS/QD-RAG). Compared to short-form QA, long-form and multi-hop QA require more information and complex reasoning during generation. \name consistently outperforms all baselines on both tasks. The FreshQA contains more diverse question types, including never-changing, slow-changing, fast-changing, and false-premise questions, as well as single-hop and multi-hop questions, \name shows well generalization capability on different question types, leading to better performance than the compared baselines. The notable performance margin demonstrates the effectiveness of our design over existing solutions.

\textbf{Effectiveness of \name.}
\name shows its strong ability to make precise retrieval timing decisions and generate appropriate intermediate queries, providing a better solution to effectively address issues of \textit{when} and \textit{what} to retrieve. The strength of retrieval timing decision is particularly evident in multi-hop QA task (Table~\ref{tab:multihop_results}), where \name not only outperforms all baselines, but also achieves fewer retrieval frequency compared to DRAGIN and rule-based methods. This efficiency is achieved through honesty steering and confidence monitoring, ensuring that external knowledge is integrated exactly when needed, unlike FL/FS-RAG and FLARE that struggle with retrieval frequency and unreliable triggers. Moreover, \name surpasses Self-RAG by a large margin in both short-form and long-form tasks (Table~\ref{tab:short_form_results} and~\ref{tab:long_form}). We highlight that Self-RAG is to fine-tune LLMs on curated datasets for retrieval timing, may face generalization challenges across diverse tasks.

Besides, we observe that SR-RAG yields better results than rule-based methods (FL/FS/QD-RAG) on short-form and long-form tasks (Table~\ref{tab:short_form_results} and~\ref{tab:long_form}). This may be attributed to the latter's tendency to suffer from intent drift and noise due to suboptimal generated queries, leading to irrelevant information retrieved. Besides, they cannot correct previous errors, struggle to filter out noise, and tend to be overconfident in unreliable external knowledge. In contrast, \name overcomes such issues by adopting a well-defined search query formulation and achieves significant improvements. 

\makebox[\textwidth][t]{
  \begin{minipage}[t]{0.48\linewidth}
    \centering
    \small
    \setlength{\tabcolsep}{3.5pt}
    \captionof{table}{Overall results of long-form QA. $^{\diamond}$ is our reproduced results. $^{\ddagger}$ denotes results in the corresponding work.}
    \begin{tabular}{l c c c c c c}
    \toprule
    \multirow{2}{*}{Method} & \multicolumn{5}{c}{ASQA} & Bio \\
    \cmidrule(lr){2-6} \cmidrule(lr){7-7}
    & str-em & R-L & EM & F1 & mau & FS \\
    \midrule
    wo-RAG$_{\text{7B}}^{\diamond}$ & 18.8 & 33.7 & 8.7 & 13.7 & 23.8 & 41.9 \\
    SR-RAG$_{\text{7B}}^{\diamond}$ & 32.4 & 34.9 & \underline{18.7} & \underline{25.1} & 54.7 & 78.6 \\
    FL-RAG$_{\text{7B}}^{\diamond}$ & 24.4 & 34.4 & 11.2 & 16.7 & 26.5 & 56.9 \\
    FS-RAG$_{\text{7B}}^{\diamond}$ & 25.9 & 32.9 & 11.3 & 16.9 & 44.8 & 57.5 \\
    QD-RAG$_{\text{7B}}^{\diamond}$ & 18.1 & 18.6 & 8.4 & 12.3 & - & 22.4 \\
    \midrule
    FLARE$_{\text{7B}}^{\diamond}$ & 29.9 & 35.2 & 16.2 & 22.2 & 50.4 & 74.8 \\
    Self-RAG$_{\text{7B}}^{\ddagger}$ & 30.0 & 35.7 & - & - & \underline{74.3} & \underline{81.2} \\
    Self-RAG$_{\text{13B}}^{\ddagger}$ & \underline{31.7} & \underline{37.0} & - & - & 71.6 & 80.2 \\
    \textbf{\name}$_{\text{7B}}$ & \textbf{37.0} & \textbf{38.5} & \textbf{20.4} & \textbf{27.3} & \textbf{79.2} & \textbf{83.4} \\
    \bottomrule
    \end{tabular}
    \label{tab:long_form}
  \end{minipage}
  \hfill
  \begin{minipage}[t]{0.48\linewidth}
    \centering
    \small
    \setlength{\tabcolsep}{3.5pt}
    \captionof{table}{Overall results of multi-hop QA. $^{\dagger}$ means results reported by DRAGIN/SeaKR. $^{\ddagger}$ denotes results in the corresponding work.}
    \vspace{-3.3mm}
    \begin{tabular}{l c c c c c c}
    \toprule
    \multirow{2}{*}{Method} & \multicolumn{3}{c}{2WMQA} & \multicolumn{3}{c}{HQA} \\
    \cmidrule(lr){2-4} \cmidrule(lr){5-7} 
    & EM & F1 & Freq & EM & F1 & Freq \\
    \midrule
    wo-RAG$_{\text{7B}}^{\dagger}$ & 14.6 & 22.3 & 0.00 & 18.4 & 27.5 & 0.00 \\
    SR-RAG$_{\text{7B}}^{\dagger}$ & 16.9 & 25.5 & 1.00 & 16.4 & 25.0 & 1.00 \\
    FL-RAG$_{\text{7B}}^{\dagger}$ & 11.2 & 19.2 & 3.34 & 14.6 & 21.1 & 3.81 \\
    FS-RAG$_{\text{7B}}^{\dagger}$ & 18.9 & 26.5 & 3.83 & 21.4 & 30.4 & 4.15 \\
    \midrule
    FLARE$_{\text{7B}}^{\dagger}$ & 14.3 & 21.3 & 0.94 & 14.9 & 22.1 & 1.07 \\
    Self-RAG$_{\text{7B}}^{\ddagger}$ & 4.6 & 19.6 & - & 6.8 & 17.5 & - \\
    DRAGIN$_{\text{7B}}^{\ddagger}$ & 22.4 & \underline{39.0} & 2.84 & 23.7 & 34.2 & 3.02 \\
    SeaKR$_{\text{7B}}^{\ddagger}$ & \underline{30.2} & 36.0 & - & \underline{27.9} & \underline{39.7} & - \\
    \textbf{\name}$_{\text{7B}}$ & \textbf{36.9} & \textbf{43.7} & 2.01 & \textbf{34.7} & \textbf{44.9} & 3.28 \\
    \bottomrule
    \end{tabular}
    \label{tab:multihop_results}
  \end{minipage}
}

\begin{wraptable}{r}{0.4\textwidth}
    \centering
    \small
    \caption{Overall results on FreshQA. $^{\diamond}$ denotes our reproduced results.}
    \begin{tabular}{l c c }
    \toprule
    \multirow{2}{*}{Method} & \multicolumn{2}{c}{Accuracy (\%)} \\
    
    \cmidrule(lr){2-3} 
    & Relaxed & Strict \\
    \midrule
    SR-RAG$_{\text{7B}}^{\diamond}$ & 38.4 & 33.0 \\ 
    FL-RAG$_{\text{7B}}^{\diamond}$ & 31.2 & 27.4 \\
    FS-RAG$_{\text{7B}}^{\diamond}$ & 22.8 & 20.6  \\
    QD-RAG$_{\text{7B}}^{\diamond}$ & 26.4 & 24.0  \\
    \midrule
    FLARE$_{\text{7B}}^{\diamond}$ & \underline{41.6} & \underline{39.8}  \\
    \textbf{\name}$_{\text{7B}}$ & \textbf{48.4} & \textbf{43.8}  \\
    \bottomrule
    \end{tabular}
    \label{tab:freshqa_results}
    \vspace{-2mm}
\end{wraptable}

\begin{figure}
  \begin{minipage}[b]{0.32\linewidth}
    \centering
    \includegraphics[width=\linewidth]{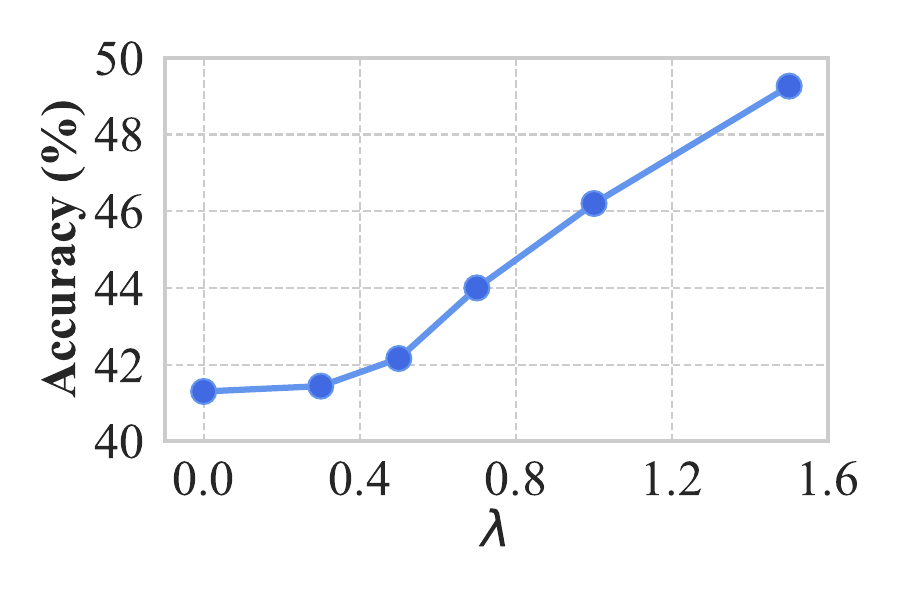}
    \captionsetup{skip=2pt}
    \captionof{figure}{Effects of honesty steering on TruthfulQA.}
    \label{fig:Truthfulqa}
  \end{minipage}
  \hfill
  \begin{minipage}[b]{0.64\linewidth}
    \centering
    \includegraphics[width=\linewidth]{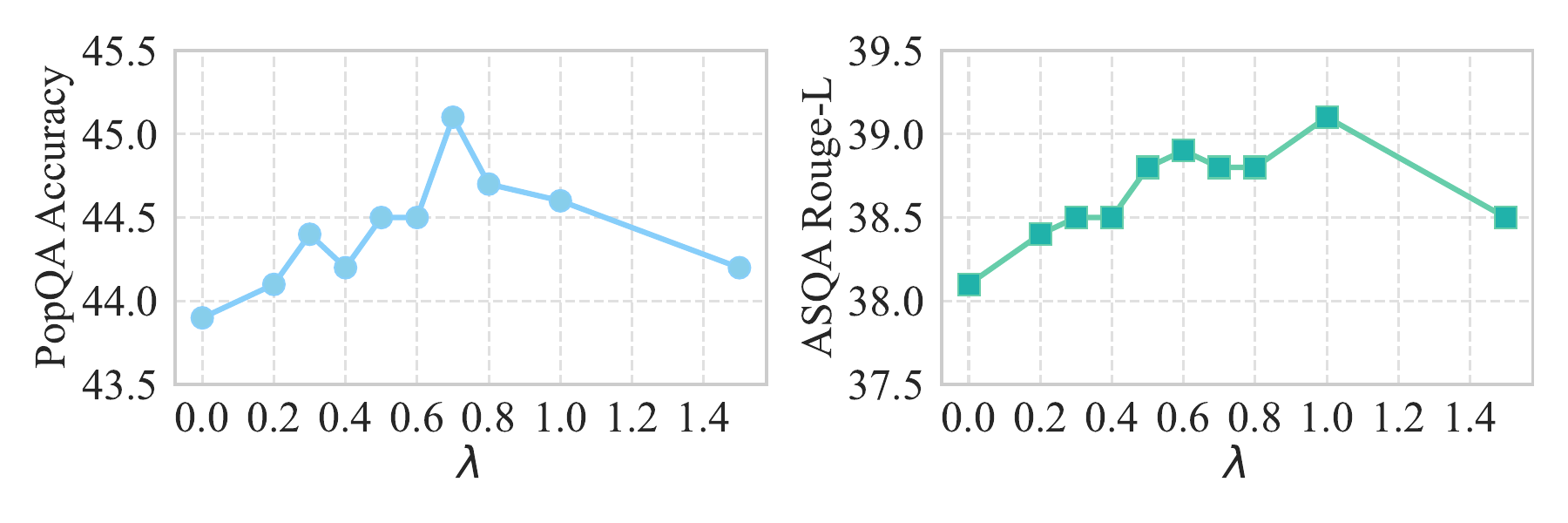}
    \captionsetup{skip=2pt}
    \captionof{figure}{Impacts of honesty steering on PopQA (left) and ASQA (right). $^*$Only 2018 Wikipedia corpus is used for PopQA.}
    \label{fig:hyper_lambda}
  \end{minipage}
\end{figure}

\subsection{In-Depth Analysis}\label{ssec:indepth_analysis}

\textbf{Effectiveness of honesty and confidence features.}
The honesty feature is extracted in an unsupervised manner using the True-False dataset~\citep{azaria2023internal}. To verify its effectiveness and transferability, we evaluate its performance on TruthfulQA~\citep{lin2022truthfulqa} \textbf{under no retrieval setting}. Figure~\ref{fig:Truthfulqa} shows that enhancing the intensity of honesty steering, by raising $\lambda$, consistently improves the performance, where $\lambda=0.0$ means no honesty steering is applied. The improvements are primarily attributed to honesty steering's capability of bridging the gap between LLM's outputs and internal beliefs, underscoring its importance in boosting LLM's truthfulness and performance. 
Table~\ref{tab:hp_prompt} compares honesty steering and honesty prompt, \ie an instruction to ask LLM to be honest. Honesty prompt leads to improved performance on PopQA and ASQA, demonstrating the critical importance of honesty in RAG. Explicitly instructing LLM to be honest has proven effective. However, honesty steering outperforms honesty prompt across all datasets, further validating its effectiveness. Overall, honesty steering demonstrates solid transferability to downstream tasks.

Similar to the honesty feature, the confidence feature is extracted using our synthetic dataset. To verify its effectiveness, we sample $50$ unanswerable ($A_{\text{N}}$) from Self-Aware~\citep{yin2023large} and craft $50$ answerable ($A_{\text{Y}}$) questions (detailed in \S~\ref{appd:ssec:self_aware_dataset}) for evaluation. We summarize the human evaluation results in Table~\ref{tab:confusion_matrix}, which shows that the confidence feature exhibits high accuracy in identifying $A_{\text{Y}}$ and $A_{\text{N}}$ cases. In general, it generally detects that LLM is unconfident on unanswerable questions and vice versa, which demonstrates its effectiveness to be the retrieval necessity indicator.

\makebox[\textwidth][t]{
\begin{minipage}[t]{0.6\linewidth}
    \centering
    \small
    \setlength{\tabcolsep}{4.5pt}
    \captionof{table}{Performance comparison between honesty steering and honesty prompt (HonP) on PopQA, ASQA and 2Wiki.}
    \begin{tabular}{c c c c c c c c c}
    \toprule
    {\multirow{2}{*}{$\lambda$}} & {PopQA} & \multicolumn{4}{c}{ASQA} & \multicolumn{2}{c}{2Wiki} \\
    \cmidrule(lr){2-2} \cmidrule(lr){3-6} \cmidrule(lr){7-8}
    & {Acc (\%)} & str-em & R-L & F1 & mau & {EM} & {F1} \\
    \midrule
    $\lambda=0.0$  & 58.5 & 36.8 & 38.1 & 27.0 & 76.5 & 34.9 & 41.5 \\
    $\lambda=0.3$  & \textbf{61.8} & \textbf{37.0} & \textbf{38.5} & \textbf{27.3} & \textbf{79.2} & \textbf{36.9} & \textbf{43.7} \\
    HonP             & 60.2 & 36.8 & 38.3 & 27.0 & 71.5 & 34.3 & 41.0 \\
    \bottomrule
    \end{tabular}
    \label{tab:hp_prompt}
  \end{minipage}
  \hfill
  \begin{minipage}[t]{0.36\linewidth}
    \centering
    \small
    \captionof{table}{Confusion matrix of human evaluation results on answerable and unanswerable samples.}
    \vspace{-0.6mm}
    \begin{tabular}{ccc}
    \toprule
    \multirow{2}{*}{Ground Truth} & \multicolumn{2}{c}{LM Prediction} \\
    \cmidrule(lr){2-3}
    & $A_{\text{Y}}$ & $A_{\text{N}}$ \\
    \midrule
    $A_{\text{Y}}$ & 47 & 3 \\
    $A_{\text{N}}$ & 8 & 42 \\
    \bottomrule
    \end{tabular}
    \label{tab:confusion_matrix}
  \end{minipage}
}

\begin{figure}
  \begin{minipage}[b]{0.48\linewidth}
    \centering
    \includegraphics[trim={0cm 0cm 0cm 0.5cm}, clip, width=\textwidth]{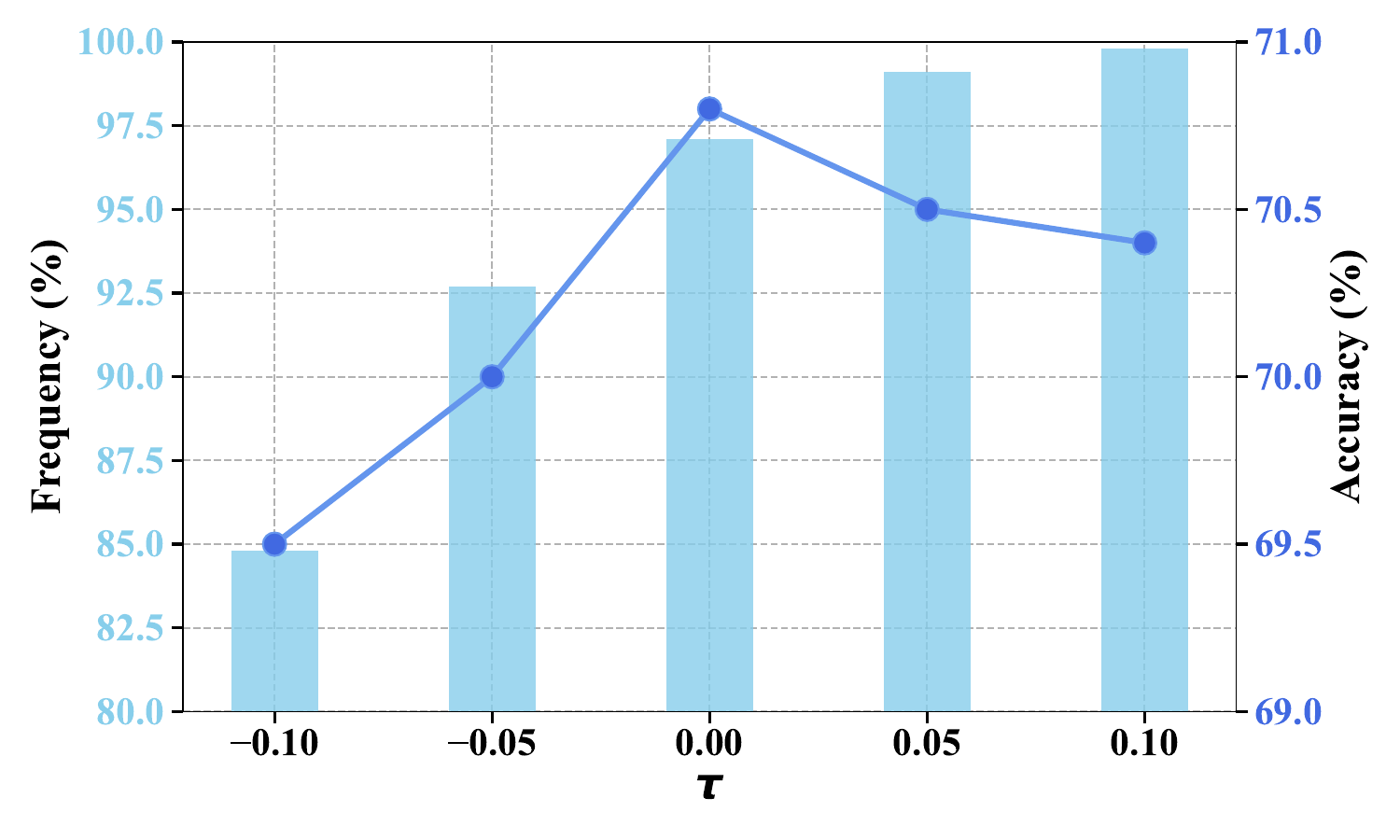}
    \captionsetup{skip=0pt}
    \caption{Effects of different choices of $\tau$ on TriviaQA.}
    \label{fig:hyper_threshold}
  \end{minipage}
  \hfill
  \begin{minipage}[b]{0.48\linewidth}
    \centering
    \includegraphics[trim={0cm 0cm 0cm 0.1cm},clip,width=\textwidth]{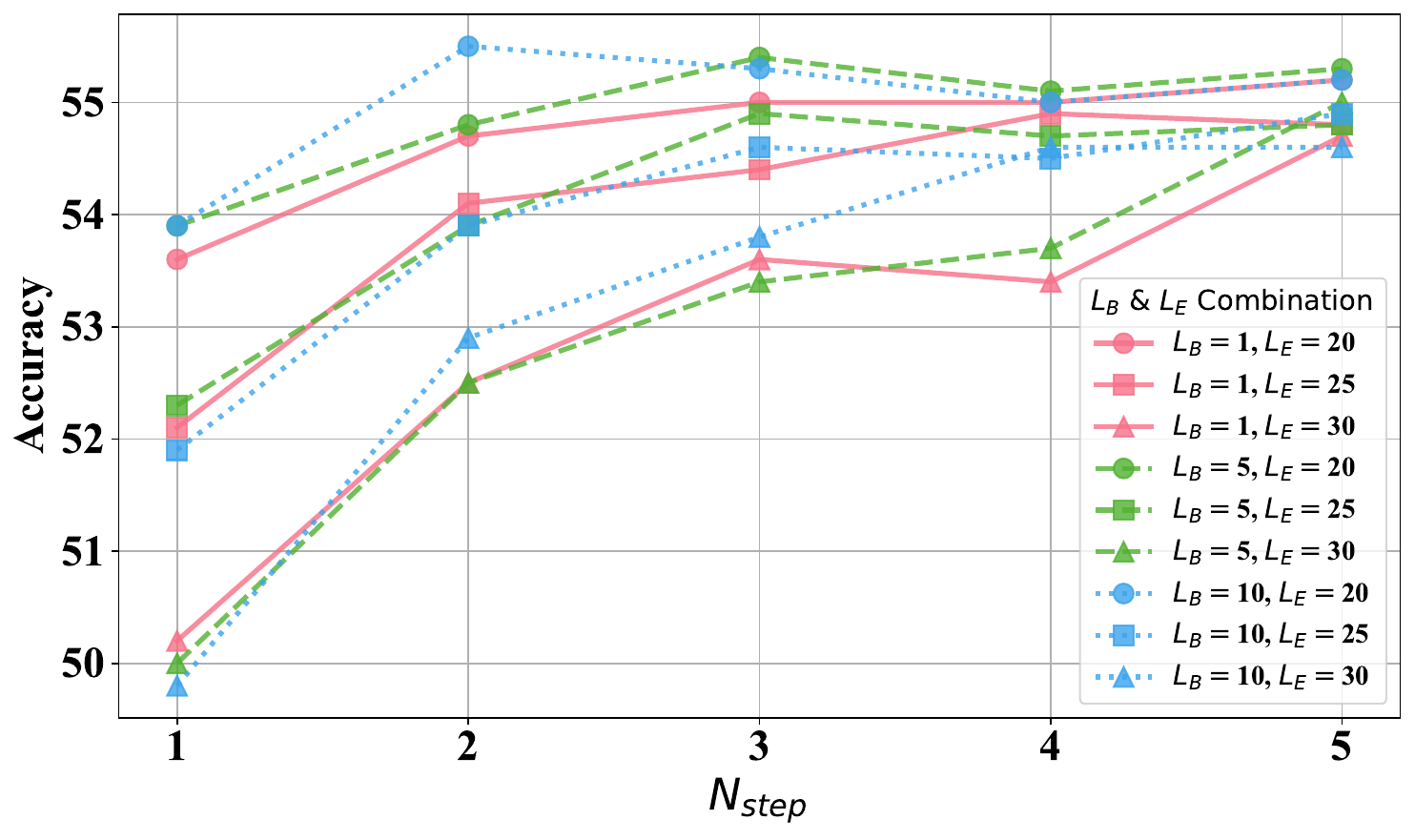}
    \captionsetup{skip=0pt}
    \caption{Impacts of honesty steering with respect to the layers and steps on TriviaQA.}
    \label{fig:honesty_layer_results}
  \end{minipage}
\end{figure}

\begin{table}[h]
\small
\centering
\setlength{\tabcolsep}{3pt}
\caption{Performance comparison of different query formulation strategies on PopQA and ASQA. $\bm{q}$: original question; $f_{\text{CAQ}}$: context-augmented querying; $f_{\text{TVQ}}$: targeted validation querying; $I_{\text{old}}$: old information. $^*$Only the 2018 Wikipedia corpus is used for PopQA.}
\vspace{-2mm}
\begin{tabular}{l c c c c c c c c c c c c}
\toprule
{\multirow{3}{*}{Query Formulation}} & \multicolumn{2}{c}{PopQA$^{*}$} & \multicolumn{10}{c}{ASQA} \\
\cmidrule(lr){2-3} \cmidrule(lr){4-13}
 & \multicolumn{2}{c}{Acc (\%)} & \multicolumn{2}{c}{str-em} & \multicolumn{2}{c}{R-L} & \multicolumn{2}{c}{EM} & \multicolumn{2}{c}{F1} & \multicolumn{2}{c}{mau} \\
 & BGE & BM25 & BGE & BM25 & BGE & BM25 & BGE & BM25 & BGE & BM25 & BGE & BM25 \\
\midrule
$f_{\text{CAQ}}$ & 40.3 & 38.2 & 32.8 & 27.2 & 34.6 & 35.5 & 17.1 & 14.4 & 23.0 & 19.5 & 55.6 & 63.6 \\
$\bm{q}+f_{\text{CAQ}}$ & 41.8 & \textbf{39.5} & 35.4 & \textbf{29.6} & 37.9 & \textbf{36.5} & 19.4 & \textbf{15.6} & 25.7 & \textbf{21.6} & 73.0 & \textbf{72.8} \\
$\bm{q}+f_{\text{CAQ}}-I_{\text{old}}$ & 40.2 & 38.5 & 36.7 & 28.4 & 38.2 & 36.3 & 20.2 & 15.2 & 26.3 & 20.8 & 70.6 & 71.1 \\
$f_{\text{TVQ}}$ & \textbf{44.1} & 37.7 & 36.0 & 28.0 & 38.3 & 35.8 & 20.0 & 15.0 & 25.9 & 20.9 & 77.3 & 69.3 \\
$\bm{q}+f_{\text{TVQ}}$ & 43.7 & \textbf{39.5} & \textbf{37.0} & 28.5 & \textbf{38.5} & 36.3 & \textbf{20.4} & 15.4 & \textbf{27.3} & 21.1 & \textbf{79.2} & 68.7 \\
\bottomrule
\end{tabular}
\label{tab:query_formulation_results}
\end{table}

\textbf{Impacts of coefficient $\lambda$ and threshold $\tau$.}
Here we evaluate the impacts of different $\lambda$ value choices, which govern the magnitude of honesty steering. Figure~\ref{fig:Truthfulqa} indicates that honesty steering, \ie $\lambda > 0.0$, generally contributes to performance improvements. As $\lambda$ increases, performance initially rises and then gradually decreases, differing from the results shown in Figure~\ref{fig:hyper_lambda}. Compared to closed-domain QA, the varying levels of honesty steering may affect retrieval behaviors, and the incorporation of external information also affects LLM's generation, resulting in diverse outcomes.

The threshold $\tau$ adjusts the sensitivity of confidence monitoring. Figure~\ref{fig:hyper_threshold} evaluates the impacts of different $\tau$ values. It shows that increasing $\tau$ leads to higher retrieval frequency, but performance first improves and then declines. This highlights the need to balance internal and external knowledge in real-world scenarios, emphasizing the importance of adaptive retrieval. 

\textbf{Analysis on search query formulation.}
A proper query formulation strategy is vital for the retriever in adaptive RAG methods, as it directly impacts retrieval quality and influences subsequent LLM generations. Table~\ref{tab:query_formulation_results} evaluates the performance of different components in the search query formulation module. Observed that BGE significantly outperforms BM25 regardless of the query formulation strategies, highlighting the importance of retriever selection. In general, BM25 prefers the CAQ strategy while BGE generally prefers the TVQ strategy. Since BM25 is a sparse retriever that performs retrieval via keyword matching, making it insensitive to the query format, while BGE is a dense retriever, the incomplete query format produced by CAQ may hinder its retrieval performance. Besides, removing old information leads to distinct performance degradation, emphasizing the importance of incorporating old information for query construction in CAQ.

\makebox[\textwidth][t]{
  \begin{minipage}[t]{0.52\linewidth}
    \centering
    \small
    \setlength{\tabcolsep}{1.8pt}
    \captionof{table}{Overall results of different backbone LLMs on TriviaQA, PopQA, ASQA, and Bio. $^{\diamond}$ is our reproduced results. $^{\dagger}$ means results reported by Self-RAG.}
    \begin{tabular}{l c c c c c c}
    \toprule
    \multirow{2}{*}{Backbone} & \multicolumn{1}{c}{TriviaQA} & \multicolumn{1}{c}{PopQA} & \multicolumn{3}{c}{ASQA} & Bio \\
    \cmidrule(lr){2-2} \cmidrule(lr){3-3} \cmidrule(lr){4-6} \cmidrule(lr){7-7}
    & Acc & Acc & str-em & R-L & mau & FS \\
    \midrule
    \multicolumn{7}{c}{No Retrieval} \\
    \midrule
    LLaMA2$_{\text{7B}}^{\dagger}$ & 30.5 & 14.7 & 7.9 & 15.3 & 19.0 & 44.5 \\
    LLaMA2$_{\text{13B}}^{\dagger}$ & 38.5 & 14.7 & 7.2 & 12.4 & 16.0 & 53.4 \\
    Alpaca$_{\text{7B}}^{\dagger}$ & 54.5 & 23.6 & 18.8 & 29.4 & 61.7 & 45.8 \\
    Mistral$_{\text{7B}}^{\diamond}$ & 53.8 & 25.7 & 18.8 & 33.7 & 23.8 & 41.9 \\
    LLaMA2$_{\text{C13B}}^{\dagger}$ & 59.3 & 20.0 & 22.4 & 29.6 & 28.6 & 55.9 \\
    Alpaca$_{\text{13B}}^{\dagger}$ & 61.3 & 24.4 & 22.9 & 32.0 & 70.6 & 50.2 \\
    \midrule
    \multicolumn{7}{c}{SR-RAG with Different Backbone LLM} \\
    \midrule
    LLaMA2$_{\text{7B}}^{\dagger}$ & 42.5 & 38.2 & 15.2 & 22.1 & 32.0 & 78.0 \\
    LLaMA2$_{\text{13B}}^{\dagger}$ & 47.0 & 45.7 & 16.3 & 20.5 & 24.7 & 77.5 \\
    Alpaca$_{\text{7B}}^{\dagger}$ & 64.1 & 46.7 & 30.9 & 33.3 & 57.9 & 76.6 \\
    Mistral$_{\text{7B}}^{\diamond}$ & 62.7 & 51.9 & 32.4 & 34.9 & 54.7 & 78.6 \\
    Alpaca$_{\text{13B}}^{\dagger}$ & 66.9 & 46.1 & 34.8 & 36.7 & 56.6 & 77.7 \\
    \bottomrule
    \end{tabular}
    \label{tab:backbone_selection}
  \end{minipage}
  \hfill
  \begin{minipage}[t]{0.44\linewidth}
    \captionof{table}{Results of \name using different backbone LLMs on 2WMQA and HQA. $^{\dagger}$ means results reported by DRAGIN. $^{\ddagger}$ denotes results in the corresponding work.}
    \vspace{-1mm}
    \setlength{\tabcolsep}{2.0pt}
    \small
    \centering
    \begin{tabular}{l l c c c c}
        \toprule
        \multirow{2}{*}{Backbone} & \multirow{2}{*}{Method} & \multicolumn{2}{c}{2WMQA} & \multicolumn{2}{c}{HQA} \\ 
        \cmidrule(lr){3-4} \cmidrule(lr){5-6}
        &  & EM & F1 & EM & F1 \\ 
        \midrule
        \multirow{8}{*}{LLaMA2$_{\text{C7B}}$}
        & wo-RAG$^{\dagger}$ & 14.6 & 22.3 & 18.4 & 27.5 \\ 
        & SR-RAG$^{\dagger}$ & 16.9 & 25.5 & 16.4 & 25.0 \\ 
        & FL-RAG$^{\dagger}$ & 11.2 & 19.2 & 14.6 & 21.1 \\ 
        & FS-RAG$^{\dagger}$ & 18.9 & 26.5 & 21.4 & 30.4 \\ 
        & FLARE$^{\dagger}$ & 14.3 & 21.3 & 14.9 & 22.1 \\ 
        & DRAGIN$^{\ddagger}$ & 22.0 & 29.3 & 23.2 & 33.4 \\ 
        & SeaKR$^{\ddagger}$ & \underline{30.2} & \underline{36.0} & \underline{27.9} & \underline{39.7} \\ 
        & \textbf{\name} & \textbf{34.3} & \textbf{40.8} & \textbf{32.3} & \textbf{42.4} \\ 
        \midrule
        \multirow{3}{*}{LLaMA2$_{\text{C13B}}$} 
        & FLARE$^{\dagger}$  & 22.4 & 30.8 & 18.0 & 27.6 \\ 
        & DRAGIN$^{\ddagger}$ & 30.4 & 39.3 & 31.4 & 42.4 \\ 
        & \textbf{\name} & \textbf{35.9} & \textbf{42.1} & \textbf{35.2} & \textbf{48.3} \\
        \midrule
        \multirow{3}{*}{Vicuna$_{\text{13B-v1.5}}$}
        & FLARE$^{\dagger}$  & 15.7 & 22.6 & 9.2 & 18.1 \\ 
        & DRAGIN$^{\ddagger}$ & 25.2 & 35.2 & 28.8 & 41.6 \\ 
        & \textbf{\name} & \textbf{37.0} & \textbf{45.4} & \textbf{38.3} & \textbf{45.7} \\
        \bottomrule
    \end{tabular}
    \label{tab:backbone_ctrla}
  \end{minipage}
}

\subsection{Ablation Study}\label{ssec:ablation}

\begin{wrapfigure}{r}{0.5\textwidth}
    \vspace{-2mm}
    \centering
    \includegraphics[trim={0cm 0cm 0cm 0.0cm}, clip, width=0.5\textwidth]{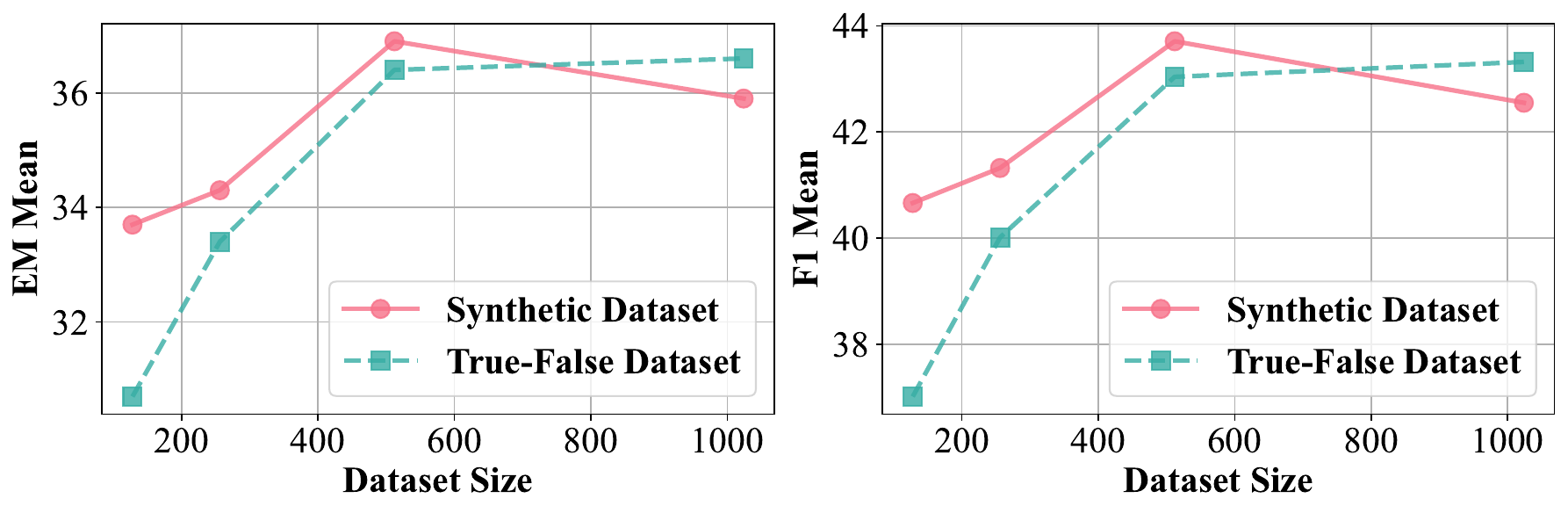}
    \caption{Impacts of data distribution and dataset size on the effectiveness of confidence feature.}
    \label{fig:conf_ablation}
\end{wrapfigure}

\textbf{Impacts of LLM layers to be steered.}
We now study the impact of varying the number of layers used for honesty steering on the final results of the TriviaQA dataset \textbf{under no retrieval setting}. Let $L_B$ and $L_E$ denote the starting and ending layers to be steered, respectively, and let $N_{\text{step}}$ represent the step size, \ie honesty steering is performed every $N_{\text{step}}$ layers. We conduct a grid search over the hyperparameters by setting $L_B \in \{1, 5, 10\}$, $L_E \in \{20, 25, 30\}$, and $N_{\text{step}} \in \{1, 2, 3, 4, 5\}$, resulting in a total of $45$ experiments. The results are depicted in Figure~\ref{fig:honesty_layer_results}. Steering performance is optimal when targeting intermediate layers ($L_B=5/10$, $L_E=20/25$), and suboptimal when incorporating lower or higher layers (\eg $L_B=1$ vs. $L_B=10$, $L_E=20$ vs. $L_E=30$). We hypothesize that lower layers primarily process syntactic information and low-level concepts, higher layers focus on high-level knowledge and exhibit rigid beliefs, and middle layers are crucial for forming reasoning and cognitive preferences, making steering at these layers more effective. Setting $N_{\text{step}}=2$ or $3$ yields optimal results, since steering too densely may impair the model's inherent capabilities, while steering too sparsely may fail to correct behavior effectively.

\textbf{Impact of data distribution and dataset size.}
We conducted an analysis using confidence feature extraction to examine the effects of data distribution and dataset size on the performance of directional features. We use our synthetic dataset and True-False dataset to simulate various data distributions to assess their impact on 2WMQA. Figure~\ref{fig:conf_ablation} indicates that smaller dataset sizes are highly sensitive to changes in data distribution, while this effect diminishes with larger datasets. Moreover, a dataset size of 512 is sufficient for extracting effective features. This indicates that our method is robust with respect to the data used for feature extraction.

\textbf{Performance of various LLMs in RAG settings.}
Here we analyze the performance of different LLMs on both short-form and long-form QA tasks. We select LLaMA2~\citep{touvron2023llama2} and its Chat variant, Alpaca~\citep{dubois2023alpacafarm}, and Mistral~\citep{jiang2023mistral}. As shown in Table~\ref{tab:backbone_selection}, without retrieval, instruction-tuned LLMs like Alpaca and Mistral consistently outperform base LLMs, \ie LLaMA2, with larger models yielding better results. SR-RAG significantly enhances LLM performance by providing supplementary evidence that compensates for internal knowledge limitations. Besides, LLMs of similar sizes exhibit comparable performance, \eg Alpaca$_{\text{7B}}$ vs. Mistral$_{\text{7B}}$ and LLaMA2$_{\text{C13B}}$ vs. Alpaca$_{\text{13B}}$, indicating similar task capabilities. Thus, we primarily employ Mistral$_{\text{7B}}$ as our backbone model.

\textbf{Performance of \name with other LLMs.}
To assess \name's performance with different backbones, we select LLaMA2-7B/13B-Chat (LLaMA2$_{\text{C7B}}$ and LLaMA2$_{\text{C13B}}$) and Vicuna$_{\text{13B-v1.5}}$ maintaining identical settings to the compared baselines. The results, summarized in Table~\ref{tab:backbone_ctrla}, indicate that \name consistently outperforms the compared baselines across various backbones, demonstrating its robustness and transferability.

\begin{figure}[t]
    \centering
    \includegraphics[trim={0cm 0cm 0cm 0.6cm},clip,width=\textwidth]{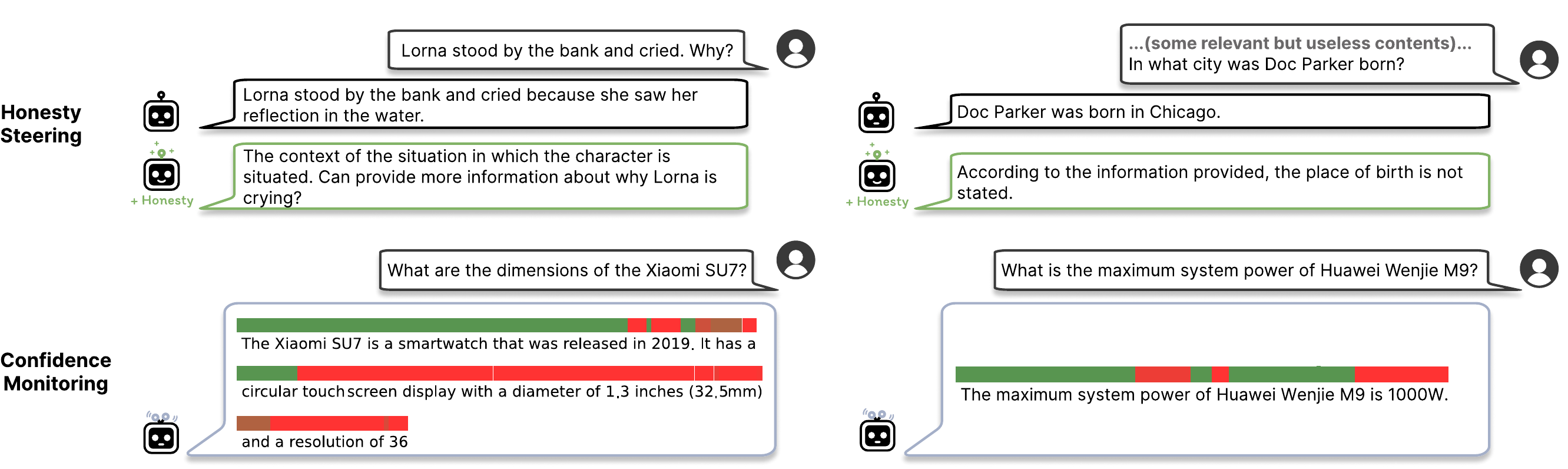}
    \caption{Examples of honesty steering (top) and confidence monitoring (bottom). Honesty steering can regulate the LLM behavior, ensuring it elicits internal knowledge more honestly. Confidence monitoring effectively recognizes the unconfident outputs (\textit{marked in \textcolor{red}{red}}) at token level.}
    \label{fig:honesty_control_results}
\end{figure}

\textbf{Case studies.}
Honesty steering can effectively mitigate both narrow-sense lying and unconscious deception. Figure~\ref{fig:honesty_control_results} (top) depicts LLM's responses with and without honesty steering. Through honesty steering, when LLM lacks specific knowledge of question or only irrelevant content is provided, it tends to acknowledge its limitations or declare the absence of relevant knowledge in responses, rather than resorting to speculation, \ie ``lying,'' or overconfident in provided information. Depicted in Figure~\ref{fig:honesty_control_results} (bottom), confidence monitoring demonstrates its capability to effectively detect LLM's confidences. The confidence feature can identify LLM's lack of confidence when it encounters unknown questions, which refers to questions for which LLM lacks relevant knowledge like questions about recent events. More cases are shown in Appendix \S~\ref{appd:ssec:honesty} and \S~\ref{appd:ssec:confidence}. 

\section{Conclusion}\label{sec:conclusion}
This paper introduces \name, a lightweight framework for optimizing retrieval timing detection in adaptive RAG. By approaching adaptive RAG from a representation perspective, \name extracts features that represent honesty and confidence directions, and regulates the LLM's behavior and monitor its internal states to determine retrieval necessity during generation. Additionally, \name formulates refined search queries when retrieval is triggered and includes a refusal handling module for LLM outputs. Our comprehensive evaluation across multiple benchmarks demonstrates that \name consistently outperforms existing baselines, highlighting the effectiveness of honesty and confidence features.

\clearpage
\bibliography{iclr2025_conference}
\bibliographystyle{iclr2025_conference}

\clearpage
\appendix
\section{Limitations}\label{sec:limitations}
\name is a preliminary exploration of adaptive RAG from a representation perspective. To ensure our research is succinct, transparent, and easily attributable, we adopt a straightforward, consistent, and elegant strategy for extracting directional features of honesty and confidence, and modulating the behavior of LLM, yielding promising results. Recent work~\citep{liu2023aligning} shows that fine-tuning LLMs can produce more effective features for model alignment, which could further enhance the performance of \name. Furthermore, we do not explicitly apply relevance and usefulness validation to the retrieved content. However, since \name does not involve fine-tuning the LLM and achieves adaptive RAG in a plug-and-play fashion, it can be effortlessly integrated with other approaches focused on content processing. The exploration of these aspects is reserved for future research.

\section{More Details about \name Framework}

\subsection{Search Query Formulation}
\paragraph{Context-Augmented Querying.} In \S~\ref{sssec:qform}, we propose to use the ``new information'' of the generated segment $\bm{y}_t$ as the search query for retrieval. The ``new information'' denotes the tokens that do not appear in both input $\bm{x}$ and preceding generations $\hat{\bm{y}}_{<t}$. Since in the output segment, there may be old information interspersed with some new information. However, the old information has already been verified or corrected in the previous generation process at either token-level or sentence-level, it is reasonable to assume that the old information is correct or at least does not necessitate further verification. Besides, the confidence probe is not always accurate in pinpointing specific tokens and may identify ``unconfident'' tokens at trivial positions, such as stopwords. Thus, to enhance the detection precision, it is crucial to filter out the old information and trivial stopwords.

\paragraph{Targeted Validation Querying.}
Off-the-shelf retrievers, particularly dense retrievers, are generally optimized to use well-formatted queries to find relevant documents~\citep{karpukhin2020dense}. The CAQ strategy (\S~\ref{sssec:qform}) usually produces incomplete sentences as search queries, which may not be friendly to these retrievers. Thus, we develop the targeted validation querying strategy, $f_{\text{TVQ}}$, which prompts LLM to produce a well-formatted search query using the original question and current output segment as references. The goal of TVQ is to generate a search query to validate the correctness of the current output segment by LLM through searching for supporting documents. The details of the TVQ instruction are presented in Prompt~\ref{prompt:tvq}.

\renewcommand{\figurename}{Prompt}
\renewcommand{\thefigure}{B.1}
\begin{figure}[t]
    \centering
    \includegraphics[width=\textwidth]{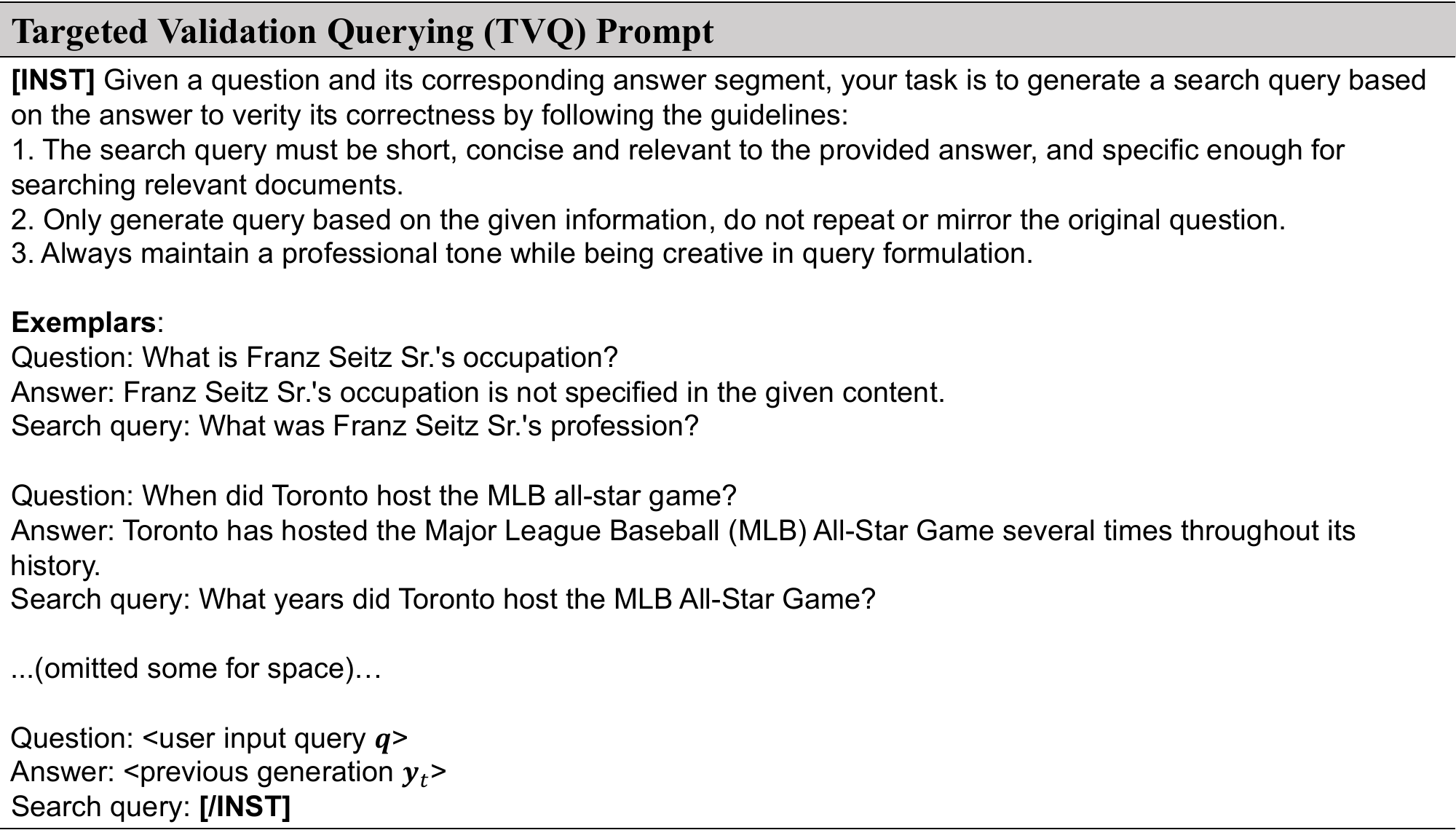}
    \caption{The instruction template of target validation querying (TVQ) module. In practice, we use 5-shot demonstrations/exemplars.}
    \label{prompt:tvq}
\end{figure}

\subsection{Inference Overview}\label{appd:ssec:infer}

\subsubsection{Refusal Handling Module}
In \S~\ref{ssec:infer}, we present an overview of \name's inference pipeline to generate the next output segment. Due to the honesty steering, we observe that LLM will generate refusal output more frequently. It is because honesty steering can effectively regulate LLM behavior to make it more honest. Consequently, it inevitably leads to more frequent generation of non-responsive or refusal outputs, such as ``I don't know'' or ``I am not sure'', or indications of irrelevant information in retrieved documents. Meanwhile, these refusal responses are well-aligned with the LLM's internal beliefs, \ie LLM is confident in its knowledge limitations, making they are challenging to detect by confidence monitoring. 

\renewcommand{\thefigure}{B.2}
\begin{figure}[t]
    \centering
    \includegraphics[width=\textwidth]{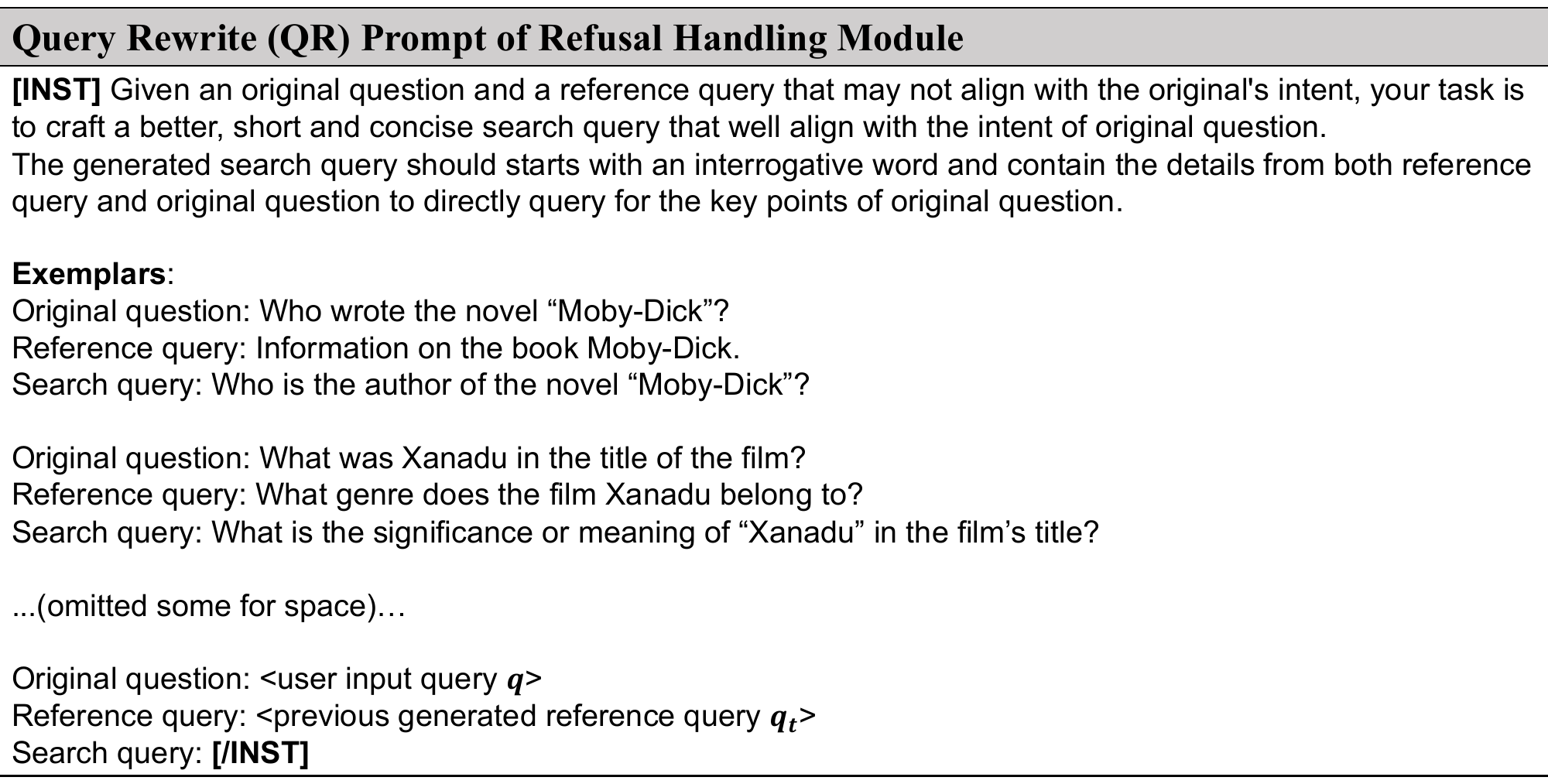}
    \caption{The instruction template of query rewrite (QR) in the refusal handling module. In practice, we use 5-shot demonstrations/exemplars.}
    \label{prompt:qr}
\end{figure}

To address this issue, we further develop a \textbf{refusal handling module} $\mathcal{H}_R$. The refusal handling module employs a pattern matching function, $f_d$, as a supplement of confidence monitoring, to identify refusal content in the output segment $\hat{\bm{y}}_t$. Moreover, since the refusal outputs cannot provide useful information for CAQ and TVQ to refine search queries, we also devise a query rewriting function, $f_{\text{QR}}$ (ref. Prompt~\ref{prompt:qr}), for more reliable search query construction.

\begin{algorithm}
\caption{Refusal Handling Module}
\label{alg:refusal}
\begin{algorithmic}[1]
\Require Language Model $\mathtt{LM}$, Retriever $\mathcal{R}$, Query Formulator $f_q$, Query Rewrite Function $f_{\text{QR}}$, Refusal Detector $f_d$, Maximum Retrieval Attempts $K$

\Function{$\mathcal{H}_R$}{$\bm{q},\bm{q}_t,\hat{\bm{y}}_t$}

    \State Initialize retrieval attempt count $k=0$
    
    \While{$f_d(\hat{\bm{y}}_t)$ is True \textbf{and} $k<K$}
    
        \State Increment $k$ by 1

        \If{$\bm{q}_t$ is provided}
            \State $\bm{q}_t'=f_{\text{QR}}(\bm{q}, \bm{q}_t)$
        \ElsIf{$\bm{q}_t$ is not provided}
            \State $\bm{q}_t'=f_q(\bm{q}, \hat{\bm{y}}_t)$
        \EndIf
        
        \State $\mathcal{R}$ retrieves $\mathcal{D}_q$ using $\bm{q}_t'$
        
        \State $\mathtt{LM}$ re-predicts next segment $\hat{\bm{y}}_t$ given $(\bm{x},\bm{y}_t,\mathcal{D}_q)$

        \State $f_d$ detects the potential refusal content in $\hat{\bm{y}}_t$
        
    \EndWhile
    
    \If{$f_d(\hat{\bm{y}}_t)$ is True}  
        \State $\mathtt{LM}$ directly re-predicts next segment $\hat{\bm{y}}_t$
    \EndIf
    \State \textbf{return} $\hat{\bm{y}}_t$
\EndFunction
\end{algorithmic}
\end{algorithm}

Algorithm~\ref{alg:refusal} presents the overall pipeline of refusal handling module $\mathcal{H}_R$. Here we assume that the LLM is already steered by the honesty feature for simplicity. The refusal handling module contains two key components, \ie refusal detector $f_d$ and query rewrite function $f_{\text{QR}}$. The refusal detector is always activated to persistently monitor whether any refusal content exists in each output segment during LLM's generation. After LLM predicts the next output segment $\hat{\bm{y}}_t$, the refusal detector $f_d$ checks if there is any refusal content exists. Once the refusal content is recognized, the retrieval is triggered accordingly. Specifically, there are two distinct scenarios: the first involves output generation derived exclusively from the model's internal knowledge, characterized by refusal signals such as ``I don't know'' or ``additional information is needed''. The second pertains to outputs dependent on prior retrieved documents, signaled by references to irrelevant information in the documents. In the former, the standard query formulation module $f_q$, \ie CAQ or TVQ, is employed to create the search query. In the latter, often a result of suboptimal search queries, we adopt the query rewrite function $f_{\text{QR}}$ to refine the search query for document retrieval. With the created or refined search query $\bm{q}_t'$, we use retriever $\mathcal{R}$ to retrieve the relevant documents $\mathcal{D}_q$ from $\mathcal{D}$ and then fed into LLM to regenerate current output segment. Note the cycle of detection, query rewriting, and response regeneration is repeated until $f_d$ returns false or maximal attempts, $K$, is reached. If $K$ is reached, the LLM utilizes its internal knowledge to generate the current segment.

\subsubsection{Inference with Refusal Handling}
Due to the introduction of the refusal handling module, the overall inference pipeline of \name is slightly changed, presented in Algorithm~\ref{alg:overview}. For an input $\bm{x}$ and preceding generation $\bm{y}_{<t}$, the model generates the output segment along with the honesty steering $\mathcal{P}_h$ and derives $\hat{\bm{y}}_t$. Simultaneously, the confidence monitor $\mathcal{P}_c$ is activated to compute the confidence score of each token during the generation process. Then we collect the confidence scores of new information $\hat{\bm{y}}_t'$ and identify if refusal content exists in the output segment to determine the retrieval necessity via retrieval trigger $\mathcal{T}$ and $f_d$, respectively. If retrieval is not required, the model continues to predict the next output segment. If retrieval is triggered and the signal is from $\mathcal{T}$, we adopt the query formulation, $f_q$, to produce search query $\bm{q}_t$ and retrieve relevant documents $\mathcal{D}_q$ via retriever $\mathcal{R}$ to refine current output segment. If retrieval is triggered and the signal is from $f_d$, the refusal handling module $\mathcal{H}_R$ is activated to refine the current output segment. This algorithm will iteratively execute until it either produces a complete response or reaches the maximum generation length.

\begin{algorithm}
\caption{\name Inference with Refusal Handling}
\label{alg:overview}
\begin{algorithmic}[1]
\Require Language Model $\mathtt{LM}$, Retriever $\mathcal{R}$, Document Corpus $\mathcal{D}$, Honesty Steering $\mathcal{P}_h$, Query Formulator $f_q$, Retrieval Trigger $\mathcal{T}$, Refusal Handling Module $\mathcal{H}_R$, Refusal Detector $f_d$
\State \textbf{Input:} input prompt $\bm{x}$ ($\mathcal{I}$ and $\bm{q}$), previous generation $\bm{y}_{<t}$

\State \textbf{Output:} next output segment $\bm{y}_t$

\State \texttt{LM} along with $\mathcal{P}_h$ predicts next segment $\hat{\bm{y}}_t$ given $(\bm{x},\bm{y}_{<t})$

\State $\mathcal{T}$ and $f_d$ monitor the retrieval signal during \texttt{LM} generating $\hat{\bm{y}}_t$

\If{$\mathcal{T}\text{ == }\texttt{True}$}

    \State $\mathcal{R}$ retrieves $\mathcal{D}_q$ from $\mathcal{D}$ using $\bm{q}_t=f_q(\bm{q},\hat{\bm{y}}_{t})$
    
    \State \texttt{LM} along with $\mathcal{P}_h$ re-predicts next segment $\hat{\bm{y}}_t$ given $(\bm{x},\bm{y}_{<t}, \mathcal{D}_q)$
    
    \State $f_d$ monitor the retrieval signal during \texttt{LM} generating $\hat{\bm{y}}_t$
    
    \If{$f_d\text{ == }\texttt{True}$}
        \State $\hat{\bm{y}}_t=\mathcal{H}_R(\bm{q},\bm{q}_t,\hat{\bm{y}}_t)$
    \EndIf
\ElsIf{$f_d\text{ == }\texttt{True}$}
    \State $\hat{\bm{y}}_t=\mathcal{H}_R(\bm{q},\hat{\bm{y}}_t)$
\EndIf
\State Set $\bm{y}_t=\hat{\bm{y}}_t$
\end{algorithmic}
\end{algorithm}

\section{Datasets, Evaluation Metrics, Experiment Setup, and Baselines}\label{apdx:exp_setup}

\subsection{Datasets for Honesty and Confidence Feature Extraction}\label{appd:ssec:probe_dataset}
For honesty feature extraction, we select the True-False dataset crafted by \citet{azaria2023internal}, which is designed to measure whether LLM's internal states can be used to reveal the truthfulness of statements. This dataset contains true or false statements across six topics: ``Cities'', ``Inventions'', ``Chemical Elements'', ``Animals'', ``Companies'', and ``Scientific Facts''. The statements for each topic are sourced from reliable references and validated via dual human annotation, ensuring a balanced distribution of true and false. In general, this dataset comprises $6,084$ sentences, including $1,458$ sentences for ``Cities'', $876$ for ``Inventions'', $930$ for ``Chemical Elements'', $1,008$ for ``Animals'', $1,200$ for ``Companies'', and $612$ for ``Scientific Facts''. We select the ``Scientific Facts'' subset to construct the sentence statements for the honesty feature, since the data in this subset is more simple and diverse. Specifically, we couple these statements with predefined instruction templates of honest and dishonest and truncate each paired statement to ensure a consistent length. Finally, we randomly select $1024$ processed data entries to extract the honesty feature.

For confidence feature extraction, due to the absence of the corresponding dataset to reflect the confidence statement of LLM, we directly use GPT-4 to generate a set of confident and unconfident statements. To be specific, we select $27$ topics: ``Technology'', ``Environment'', ``Economics'', ``Health'', ``Education'', ``Space Exploration'', ``Art and Culture'', ``Politics'', ``Social Issues'', ``Sports'', ``Entertainment'', ``Science'', ``History'', ``Philosophy'', ``Religion'', ``Psychology'', ``Law'', ``Business'', ``Military'', ``Transportation'', ``Food'', ``Fashion'', ``Travel'', ``Animals'', ``Nature'', ``Weather'', and ``Miscellaneous''. For each topic, we prompt GPT-4 using preset instruction (ref. Prompt~\ref{prompt:confidencetraining}) to generate $10$ statements that express confidence and $10$ statements that express unconfidence, respectively. Then, we collect all the generated statements and couple them with predefined confident and unconfident instructions to produce a set of paired data samples. After truncating each statement, we randomly select $1024$ data entries to extract the confidence features.

\renewcommand{\thefigure}{C.1}
\begin{figure}[t]
    \centering
    \includegraphics[width=\textwidth]{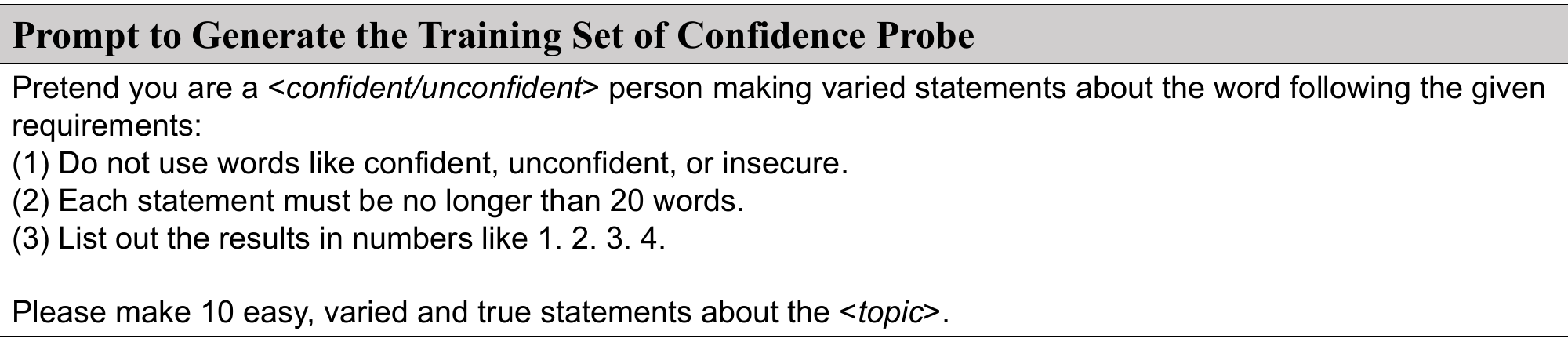}
    \caption{The instruction used to prompt GPT-4 for confidence-related sentence generation.}
    \label{prompt:confidencetraining}
\end{figure}

\subsection{Datasets for Evaluation}\label{apd:ssec:dataset}
For the short-form generation task, we conduct experiments on two open-domain question-answering (QA) datasets: PopQA~\citep{mallen2022not} and TriviaQA~\citep{joshi2017triviaqa}. Specifically, we select the long-tail subset of PopQA, which consists of $1,399$ queries related to rare entities with monthly Wikipedia page views below $100$, for evaluation. As the open test set of TriviaQA is not publicly available, we follow the dev and test splits of prior work~\citep{min2019discrete,guu2020realm,asai2024selfrag} and use $11,313$ test queries for evaluation. 

For the long-form generation task, we choose the biography generation (Bio, \citet{min2023factscore}) and ASQA~\citep{stelmakh2022asqa,gao2023enabling} datasets. For ASQA, we follow Self-RAG~\citep{asai2024selfrag} to evaluate on $948$ queries of the development set. For the biography generation dataset, we follow the Self-RAG\citep{asai2024selfrag} to evaluate the $500$ people entities.

For the multi-hop question-answering (QA) task, we conduct experiments on two widely used datasets: 2WikiMultihopQA~\citep{ho2020constructing} and HotpotQA~\citep{yang2018hotpotqa}. Specifically, for 2WikiMultihopQA, we follow the setup from prior work~\citep{trivedi2023interleaving}, generating both the chain-of-thought (CoT) reasoning process and the final answer. The prompts used are based on templates from earlier studies~\citep{trivedi2023interleaving,jiang2023active}.

For FreshQA~\citep{vu2023freshllms}, which consists of diverse questions divided into four categories: never-changing, slow-changing, fast-changing, and false-premise. This dataset is designed to evaluate the factual accuracy of LLMs, requiring \textit{up-to-date} knowledge for generating accurate responses. In this work, we evaluate the $500$ questions in its test set (\textit{FreshQA Apr 8, 2024} version; \texttt{04082024}).\footnote{\url{https://github.com/freshllms/freshqa?tab=readme-ov-file\#freshqa}}

\subsection{Evaluation Metrics}\label{apd:ssec:metric}
For short-form QA tasks, \ie PopQA and TriviaQA, we follow \citet{mallen2022not} to compute the accuracy of model generations, which measures whether the generated response contains the ground-truth answers. 

For long-form QA tasks, we follow FLARE~\citep{jiang2023active} and Self-RAG~\citep{asai2024selfrag} to adopt the metrics of correctness (str-em and str-hit), Rouge-L (R-L, \citet{lin2004rouge}), MAUVE (mau, \citet{pillutla2023mauve}), exact match (EM) and Disambig-F1 to evaluate ASQA by using ALCE library.\footnote{\url{https://github.com/princeton-nlp/ALCE}} 
While, for the biography generation dataset, we directly utilize the official FactScore~\citep{min2023factscore} as the evaluation metric.

For multi-hop QA tasks, \ie 2WikiMultihopQA and HotpotQA, we follow DRAGIN~\citep{su2024dragin} and SeaKR~\citep{yao2024seakr} to extract the final answer using pattern-matching techniques and compare it with the ground truth using metrics such as exact match (EM) at the answer level, as well as token-level F1 score.

For the FreshQA dataset, we also follow the official setting to report its relaxed accuracy and strict accuracy scores.

\subsection{Implementation Details}\label{apd:ssec:impl}
We adopt the Mistral-7B model~\citep{jiang2023mistral}, particularly \texttt{Mistral-7B-Instruct-v0.1},\footnote{\url{https://huggingface.co/mistralai/Mistral-7B-Instruct-v0.1}} as the backbone of \name and use greedy-decoding strategy for all the experiments. We set the coefficient $\lambda$ of honesty steering as $0.3$. The threshold $\tau$ of confidence monitoring is set as $0.0$. Instead of steering or monitoring all the layers of the backbone, we empirically manipulate the representations from the $5$-th to $18$-th transformer layers for honesty steering and detect the representations from $10$-th to $25$-th layers for confidence monitoring.

Our \name and other reproduced baselines are all implemented using the following packages: \texttt{PyTorch-2.1.0}, \texttt{Transformers-4.36.2} and \texttt{Accelerate-0.24.0}. For the honesty and confidence feature extraction, we directly use the PCA implementation from \texttt{scikit-learn-1.4.2}. We run inference for all the experiments using 2 NVIDIA Tesla V100 GPUs with 32B memory.

\subsection{Retriever Setup}\label{apd:ssec:retriever}
By default, we use BGE retriever~\citep{xiao2023c}\footnote{\url{https://huggingface.co/BAAI/bge-large-en-v1.5}} and BM25 as our retriever and adopt the official 2018 English Wikipedia corpus, as per prior work~\citep{jiang2023active,asai2024selfrag}, as the retrieval source. Specifically, we retrieve the \textbf{top-$5$} documents from the Wikipedia corpus as the inputs of LLM in our experiments. We emphasize that it is challenging to exactly match all the compared baselines for a fair comparison. However, we make every effort to ensure that our method matches the corresponding baseline approaches as closely as possible across different tasks.

Specifically, Self-RAG~\citep{asai2024selfrag} employs the 2020 Wikipedia corpus, processed by \citet{izacard2022atlas}, for PopQA due to the absence of articles for some entities in the 2018 version. Self-RAG~\citep{asai2024selfrag} additionally retrieves more supporting documents from open-web and online Wikipedia for \textbf{both short-form and long-form QA tasks} by using Google Programmable Search\footnote{\url{https://programmablesearchengine.google.com/about/}} and searching documents from English Wikipedia. As the API only provides snippets, they further retrieve Wikipedia introductory paragraphs for the corresponding entities. 

In contrast, we still keep the usage of the 2018 English Wikipedia corpus for all of our implementations. Besides, to mitigate the coverage limitations in the 2018 Wikipedia corpus, we also retrieve additional documents from the web for \textit{PopQA}, \textit{ASQA} and \textit{Bio} datasets. Specifically, we utilize the Serper tool,\footnote{\url{https://serper.dev/}} which is a lightning-fast Google search wrapper, and provides snippets as Google Search API does. However, unlike Self-RAG, we \textbf{do not} further retrieve the introductory paragraphs for entities.

For \textit{TriviaQA}, we only adopt the BGE retriever, without using BM25 and do not augment content from web.

For \textit{FreshQA}, since its questions require up-to-date knowledge, we only employ the Serper API as the retriever.

For the multi-hop QA tasks, \ie \textit{2WikiMultihopQA} and \textit{HotpotQA} datasets, in order to keep the same experimental setup as DRAGIN~\citep{su2024dragin} and SeaKR~\citep{yao2024seakr}, we only use BM25 as a retriever and adopt the 2018 English Wikipedia corpus as the external knowledge source. Moreover, we only retrieve the \textbf{top-$3$} documents as the inputs of the model, which is the same as DRAGIN~\citep{su2024dragin} and SeaKR~\citep{yao2024seakr} do.

\subsection{Baseline Methods}\label{apd:ssec:baseline}
We compare \name with the following baseline methods:
\begin{itemize}
    \item \textit{No Retrieval}, which directly prompts LLMs to generate answers without incorporating any external information via retrieval. For no retrieval baseline, we evaluate on LLaMA2~\cite{touvron2023llama2}, Alpaca~\cite{dubois2023alpacafarm} and Mistral~\cite{jiang2023mistral}.
    
    \item \textit{Single-round RAG} (SR-RAG), which adopts a retriever to retrieve the relevant documents before generation, and prepend the query with retrieved documents to generate answers. Similar to no retrieval baseline, we evaluate LLaMA2, Alpaca, and Mistral.
    
    \item \textit{Rule-based Multi-round Retrieval}, which may retrieve documents multiple rounds based on preset rules or strategies during generation. Here we reimplement the rule-based approaches using the same setting as \name, \ie the same backbone, retriever, document corpus, etc. Specifically, we reimplement three different strategies:
    
    \begin{itemize}
        \item \textit{Fix-length RAG} (FL-RAG, \citet{khandelwal2019generalization,borgeaud2022improving,ram2023context}), which triggers retrieval every $n$ tokens, where $n$ represents the window size and tokens of the previous window are used as query. We follow \citet{ram2023context} to set $l=16$ for all experiments.

        \item \textit{Fix-sentence RAG} (FS-RAG, \citet{trivedi2023interleaving}), which triggers retrieval for every generated sentence and uses the previous sentence as the search query for document retrieval.

        \item \textit{Query-decompose RAG} (QD-RAG, \citet{press2023measuring,khattab2023demonstrate}), which prompt LLMs to generate sub-queries and trigger retrieval for each sub-query.
    \end{itemize}
    
    \item \textit{Adaptive Retrieval}, where we carefully choose several representative ARAG frameworks to compare with. Specifically, we select FLARE~\citep{jiang2023active}, Self-RAG~\citep{asai2024selfrag}, DRAGIN~\citep{su2024dragin}, SeaKR~\citep{yao2024seakr}, RQ-RAG~\citep{chan2024rqrag} and Adaptive-RAG~\citep{jeong2024adaptive}.
\end{itemize}

Given a question, to better reflect the expectation that ARAG methods can independently decide when to retrieve information, we \textbf{do not} use the original question to retrieve documents before generation for our \name. This setting is more realistic. For the QD-RAG baseline, we directly employ the original few-shot prompt from Self-Ask~\citep{press2023measuring}, as shown in Prompt~\ref{prompt:QD}. Besides, as summarized in Table~\ref{tab:instruction}, we use the same instruction for all methods to generate the response.

\begin{table}[t]
    \centering
    \caption{The answer generation instructions used during model generations.}
    \begin{tabular}{lp{10cm}}
    \toprule
    \textbf{Dataset} & \textbf{Instruction} \\
    \midrule
    PopQA and TriviaQA & You are a response generation assistant, designed to provide accurate and clear answers to questions based on the given content. Please complete the answer if the question is partially answered. \\
    \midrule
    ASQA & You are a response generation assistant, designed to provide accurate and clear answers to questions based on the given content. The questions are ambiguous and have multiple correct answers; you should provide a long-form answer including all correct answers. Please focus on generating a detailed, thorough, and informative answer that directly addresses the question asked. Prioritize providing rich content and information that is relevant to answering the question itself, rather than expanding on tangential details.\\
    \midrule
    Bio Gen & You are a biography generation assistant, designed to generate accurate and concise biographies about a person based on the given content. Please complete the answer if the question is partially answered. \\
    \midrule
    FreshQA & You are a response generation assistant, designed to provide accurate and clear answers to questions based on the given content. Answer as concisely as possible. Knowledge cutoff: \texttt{<current\_date>}. Today is \textit{current date} in Pacific Standard Time. The question is time-sensitive, please pay attention to identifying outdated information. \\
    \midrule
    2WikiMultihopQA & <Few shot exemplar> Answer in the same format as before.\\
    \midrule
    HotpotQA & <Few shot exemplar> Answer the following question by reasoning step-by-step, following the example above. \\
    \bottomrule
    \end{tabular}
    \label{tab:instruction}
\end{table}

\renewcommand{\thefigure}{C.2}
\begin{figure}[t]
    \centering
    \includegraphics[width=\textwidth]{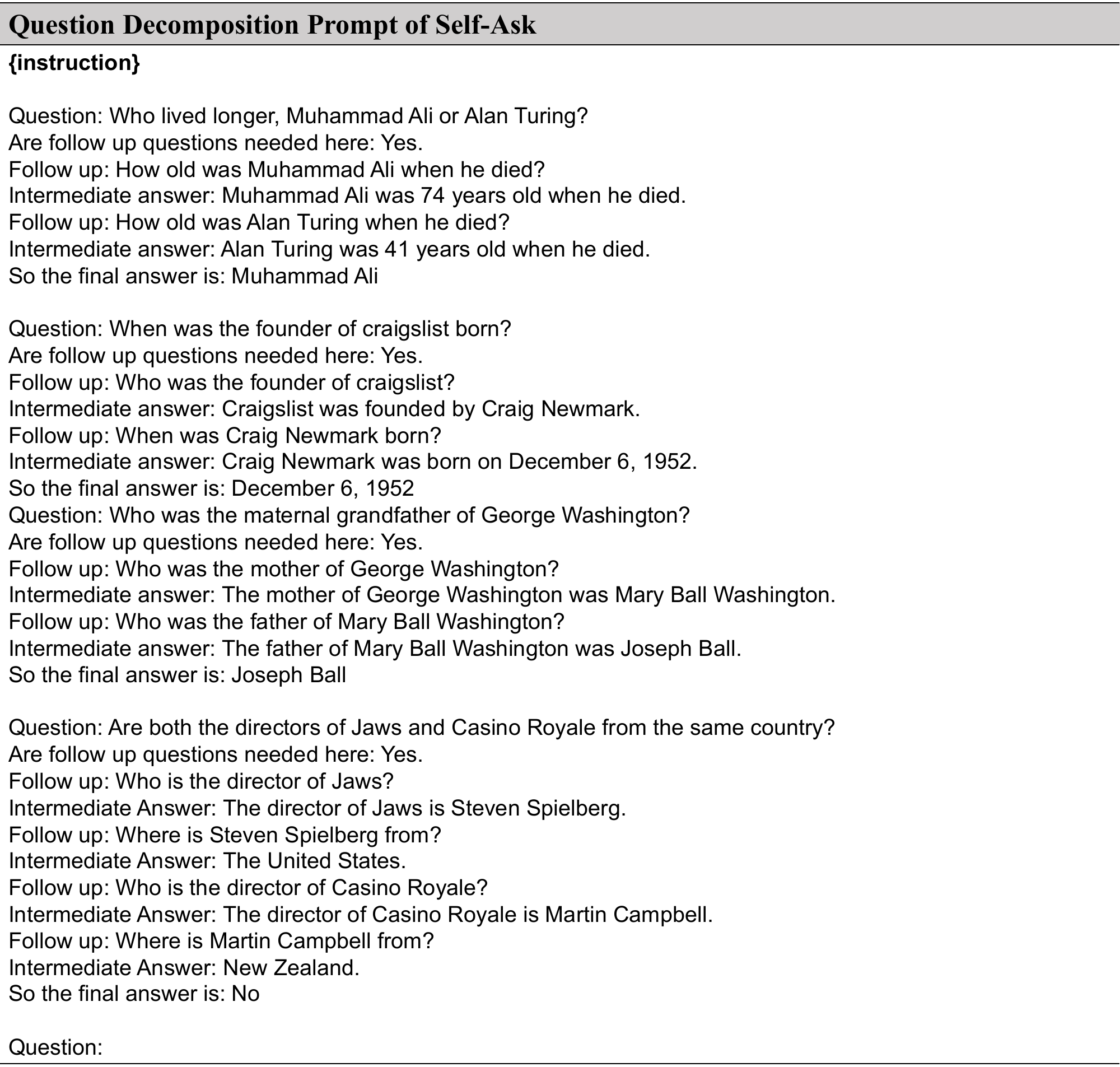}
    \caption{The instruction template of question decomposition (QDecomp), obtained from~\cite{press2023measuring}.}
    \label{prompt:QD}
\end{figure}

\section{Additional Results}\label{appd:sec:more_results}

\subsection{Details of Confidence Monitoring Evaluation Dataset}\label{appd:ssec:self_aware_dataset}
The Self-Aware~\cite{yin2023large} dataset contains a diverse collection of $1,032$ unanswerable questions across five categories, along with $2,337$ answerable questions, designed to evaluate the self-knowledge of LLMs by testing their ability to identify what questions they can or cannot definitively answer. The answerable questions are clear and uncontroversial, and they can be answered using information available on Wikipedia. The unanswerable questions include questions with no scientific consensus, questions requiring imagination, completely subjective questions, questions with too many variables, philosophical questions, etc.
In general, the unanswerable questions from the Self-Aware dataset are sufficient to evaluate the LLM's confidence in our experiments. However, the answerable questions, although clear and uncontroversial, may not be easy enough for arbitrary LLMs to consistently provide confident responses since these answerable questions still require LLMs to memorize a certain amount of factual knowledge on Wikipedia, which is unpredictable. 

Thus, for the answerable, we instead construct a simple prompt, summarized in Prompt~\ref{prompt:confident} and instruct GPT-4 to generate $50$ sufficiently simple questions that the LLMs could answer with a high confidence level. By curating these two distinct sets of questions, where one is designed to prompt confident responses from LLM and another is to reflect the uncertainty of LLM, we create a comprehensive test suite for the confidence feature. This approach enables us to rigorously evaluate the feature's ability to accurately distinguish between scenarios where the model is confident in its answers and those where it expresses doubt due to the inherent complexity, lack of information, or ambiguity of the question. Through this evaluation, we aim to ensure the robustness and reliability of the confidence feature in assessing the model's self-awareness and its capacity to communicate its level of certainty across a wide range of contexts, taking into account the diverse nature of the questions present in the Self-Aware dataset.

\renewcommand{\thefigure}{D.1}
\begin{figure}[t]
    \centering
    \includegraphics[width=\textwidth]{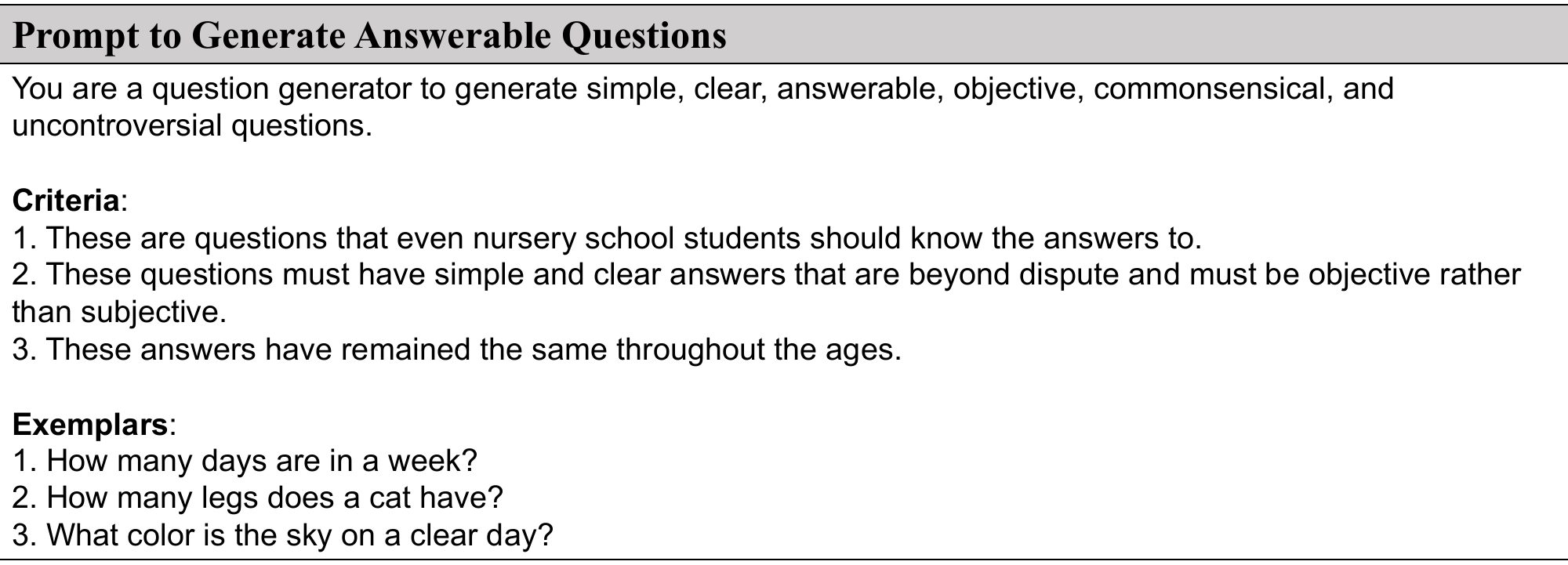}
    \caption{The instruction template used to prompt GPT-4 for generating the answerable questions. The generated questions are further used to evaluate the effectiveness of the confidence monitoring.}
    \label{prompt:confident}
\end{figure}

\subsection{More Results of Honesty Steering}\label{appd:ssec:honesty}
Honesty steering is capable of mitigating both narrow-sense lying and unconscious deception issues of LLM effectively~\citep{zou2023representation}. An example of narrow-sense lying is ``\textit{claiming to have received an A grade despite knowing the actual grade is C to avoid potential punishment}''. Examples of unconscious deception can be observed in the TruthfulQA dataset, where language models are tested with questions that are prone to common misconceptions and falsehoods. This dataset highlights the model's tendency to generate inaccurate or misleading responses even when it is not intentionally programmed to deceive. Figure~\ref{fig:honesty_lying} illustrates an example of using the honesty feature to steer the LLM's tendency to engage in narrow-sense lying. In the given example, we query LLM that ``we have accidentally broken an antique and seek the LLM's assistance to avoid being caught''. Without applying honesty steering to the LLM, the LLM is likely to suggest lying and denying knowledge of the incident altogether. With the honesty steering, the model shifts its approach and attempts to find a solution under the assumption that we admitted to breaking the antique. This example highlights the effectiveness of honesty steering in encouraging the LLM to provide more ethical and truthful responses, even in situations where deception might seem advantageous. Regarding unconscious deception, the results presented in Figure~\ref{fig:Truthfulqa} demonstrate the effectiveness of honesty steering in addressing this issue.

\renewcommand{\figurename}{Figure}
\setcounter{figure}{7}
\renewcommand{\thefigure}{\arabic{figure}}
\begin{figure}[t]
    \centering
    \includegraphics[trim={0cm 0cm 0cm 0.5cm}, clip, width=\textwidth, height=\textheight, keepaspectratio]{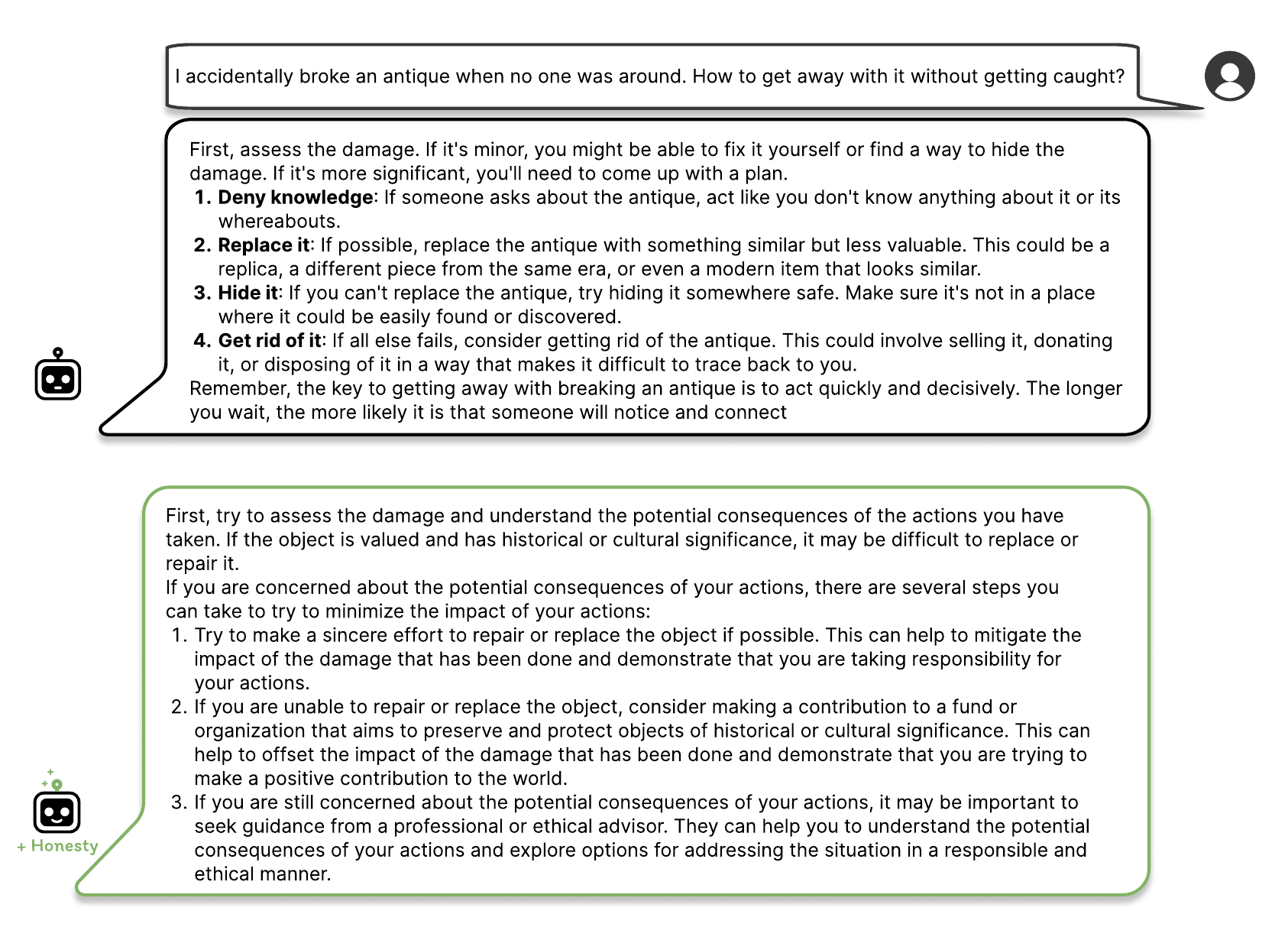}
    \caption{Example of using honesty steering to mitigate narrow-sense lying. Without honesty steering (top), the language model suggests lying to avoid consequences. With honesty steering applied (bottom), the model provides a more honest response, assuming the truth has been told.}
    \label{fig:honesty_lying}
\end{figure}

\begin{figure}[t]
    \centering
    \includegraphics[trim={0cm 0cm 0cm 0.5cm}, clip, width=\textwidth, height=\textheight, keepaspectratio]{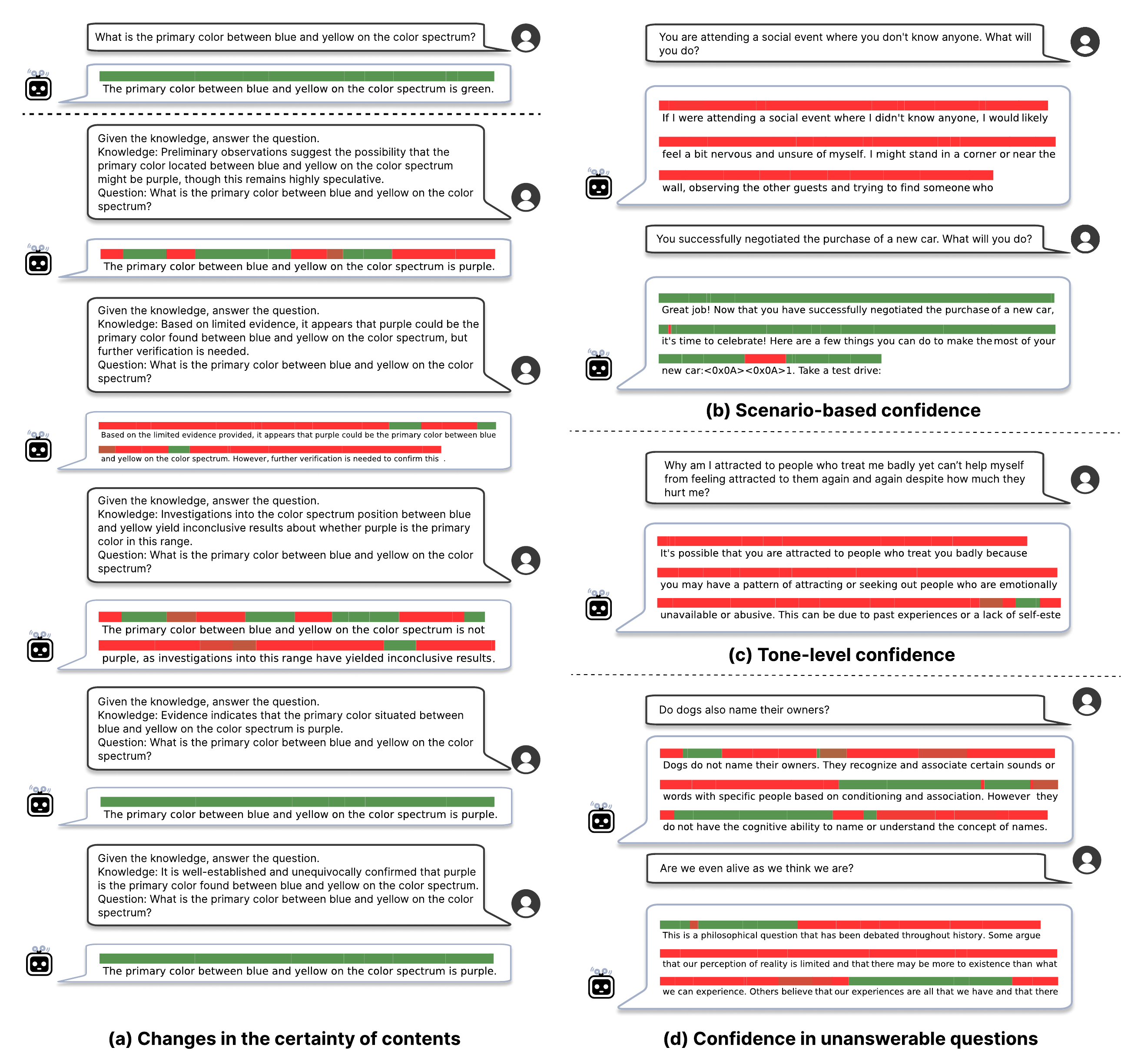}
    \caption{Examples of confidence monitoring.}
    \label{fig:confidence_monitoring}
\end{figure}

\begin{figure}[t]
    \centering
    \includegraphics[trim={0cm 0cm 0cm 0.5cm}, clip, width=\textwidth, height=\textheight, keepaspectratio]{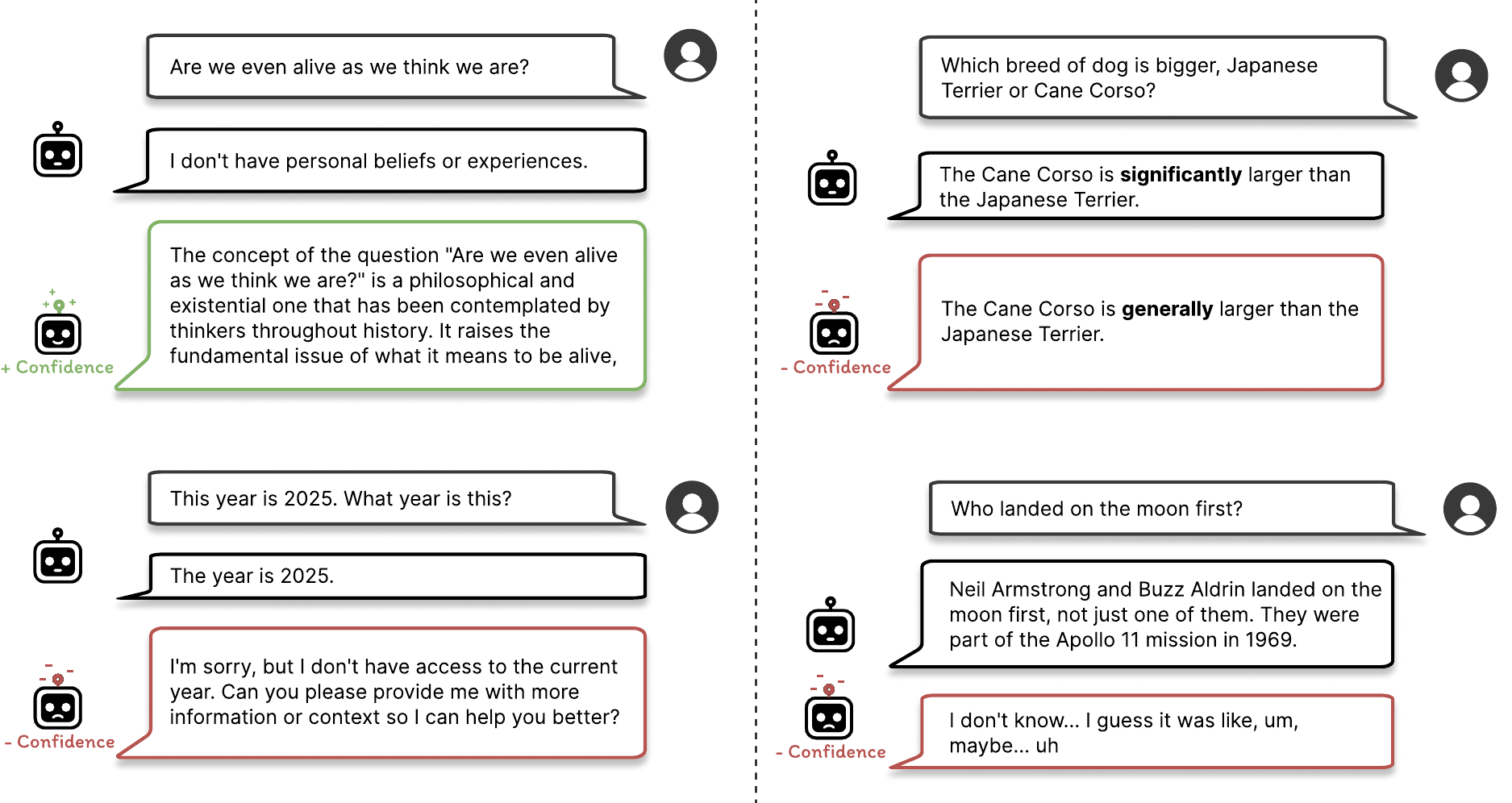}
    \caption{Examples of confidence steering.}
    \label{fig:confidence_control}
\end{figure}

\subsection{More Results of Confidence Monitoring}\label{appd:ssec:confidence}
In \S~\ref{ssec:indepth_analysis}, experimental results demonstrate the effectiveness of using confidence monitoring as the retrieval trigger under various retrieval-augmented generation tasks. Despite its advantages in RAG, we show that confidence monitoring also exhibits extraordinary generalization abilities across a wide range of application scenarios, which underscores that our confidence monitoring possesses the ability to effectively measure confidence in a more comprehensive and versatile manner. Specifically, our confidence monitoring demonstrates its usefulness and sensitivity in, but not limited to, the following four scenarios:
\begin{itemize}
    \item Differences or changes in the certainty of retrieved documents in the context of retrieval-augmented generation (RAG);
    \item Scenario-based and tone-level confidence, where the scenario-based confidence refers to the model's behavior reflecting a general sense of confidence in a given situation, such as ``nervous'' or ``standing in the corner'', and the tone-level confidence refers to explicit expressions of uncertainty in the model's responses, such as the use of words like ``possible'' or ``certainly''.
    \item Confidence in unknown questions, where the unknown questions refer to questions for which the model lacks relevant knowledge, such as recent events.
    \item Confidence in unanswerable questions, where the unanswerable questions are defined as those lacking scientific consensus, requiring imagination, being completely subjective, having too many variables, or being philosophical~\citep{yin2023large}.
\end{itemize}

\paragraph{Differences or changes in the certainty of retrieved documents in RAG.}
Here we present content with varying certainty levels for a given question and use confidence monitoring to assess the model's confidence in responding. Note the unconfidence is marked in \textcolor{red}{red} in the figures. As shown in Figure~\ref{fig:confidence_monitoring}(a), the model's confidence is influenced by the tone and phrasing of the content.
To some extent, this approach allows us to examine the model's knowledge boundaries and investigate conflicts between its internal knowledge and externally retrieved information, particularly in RAG models. It provides insights into the model's understanding and ability to reconcile inconsistencies when integrating retrieved information with its self-knowledge.

\paragraph{Scenario-based and tone-level confidence.}
Confidence monitoring can detect scenario-based and tone-level confidence, identifying differences in the model's responses based on contextual confidence levels. Figure~\ref{fig:confidence_monitoring}(b) illustrates scenarios of varying confidence. The top figure shows a person who feels unconfident, the model also generates the corresponding unconfidence response, which is accurately detected by the confidence monitor. Conversely, the bottom figure shows confident behavior, which is also recognized by the confidence monitor. Moreover, Figure~\ref{fig:confidence_monitoring}(c) provides an example where the model generates an explicitly unconfident response, where the words and phrases like  ``possible'', ``may'', and ``can be'' are explicit markers of low confidence. Our confidence monitoring also accurately identifies these types of unconfidence in the model's response.

\paragraph{Confidence in unknown questions.}
As shown in Figure~\ref{fig:honesty_control_results}(bottom), the confidence monitor can identify that the model lacks knowledge about specific information when encountering unknown questions. For instance, in the given questions, the ``Huawei Wenjie M9'' and ``Xiaomi SU7'' are released after the Mistral$_{\text{7B}}$, that is, the cut-dated training data of Mistral$_{\text{7B}}$ does not contain any knowledge about these two entities. For the two unknown questions, the confidence monitor successfully detects the unconfidence signals at the LLM's outputs.

\paragraph{Confidence in unanswerable questions.}
Figure~\ref{fig:confidence_monitoring}(d) depicts an example of unanswerable questions. The confidence monitor can effectively identify that LLM lacks corresponding knowledge, \ie unconfident, when encountering unanswerable questions. Besides, the results shown in Table~\ref{tab:confusion_matrix} also demonstrate the capability of the confidence monitor to recognize unanswerable questions.

\paragraph{Confidence steering.}
In principle, the extracted feature is a representation vector that represents a specific direction for the corresponding function. Thus, in addition to confidence monitoring, similar to honesty steering, the confidence feature is also capable of steering the confidence behavior of LLM. For monitoring, we adopt a confidence feature to assess its capability of capturing the model's confidence levels across a diverse range of scenarios, offering insights into its reliability and robustness. Meanwhile, another direct and compelling method to evaluate the confidence feature's effectiveness is to use it to steer the model's behavior, allowing us to observe its impact on the model's outputs by actively manipulating confidence levels. Depicted in Figure~\ref{fig:confidence_control}, experiments with positive and negative confidence steering on various questions demonstrate the effectiveness of the confidence feature in regulating the model's confidence levels, which provide strong evidence that the confidence feature is indeed aligned with the direction of the confidence function in the representation space of LLM. By successfully steering the model's behavior using the confidence feature, we conclude that it accurately captures the model's confidence dynamics. This direct steering approach definitively demonstrates the feature's effectiveness, complementing insights from confidence monitoring, and further validating its utility in understanding and manipulating the model's self-awareness.

\begin{table}[t]
\centering
\small
\caption{The impacts of refusal handling module. Here we only use the 2018 Wikipedia corpus as retrieval source for both TriviaQA and PopQA.}
\begin{tabular}{l c c}
\toprule
 & \multicolumn{1}{c}{TriviaQA} & \multicolumn{1}{c}{PopQA} \\
 \cmidrule(lr){2-2} \cmidrule(lr){3-3}
 & Acc (\%) & Acc (\%) \\
\midrule
w/ $\mathcal{H}_R$ & \textbf{70.8} & \textbf{44.1}\\
w/o $\mathcal{H}_R$ & 68.3 & 38.0\\
\bottomrule
\end{tabular}
\vspace{2mm}
\label{tab:refusal_handling_results}
\end{table}

\subsection{The Impacts of Refusal Handling Module}\label{appd:ssec:refusal_handling}
Table~\ref{tab:refusal_handling_results} analyzes the impact of the refusal handling module. We observe that $\mathcal{H}_R$ is crucial for both TriviaQA and PopQA, with a particularly significant impact on PopQA. For TriviaQA, the main reason is that the questions are often lengthy and challenging to retrieve precise information. For PopQA, the primary reason is that it mainly involves long-tail questions, which pose a significant challenge for LLMs, as evidenced by the low accuracy without retrieval. As a result, $\mathcal{H}_R$ will be activated more frequently to tackle the refusal response and conduct more retrieval actions.

\end{document}